%% file: acl_latex.tex
\pdfoutput=1

\documentclass[11pt]{article}

\usepackage{times}
\usepackage{latexsym}

\usepackage[T1]{fontenc}

\usepackage[utf8]{inputenc}

\usepackage{microtype}

\usepackage{inconsolata}

\usepackage{microtype}
\usepackage{arydshln}
\PassOptionsToPackage{table}{xcolor}
\usepackage[preprint]{acl}
\usepackage{graphicx}
\usepackage{subfigure}
\usepackage{enumitem}
\usepackage{booktabs} 

\usepackage{amsmath}
\usepackage{amssymb}
\usepackage{bbm}
\usepackage{mathtools}
\usepackage{amsthm}
\usepackage{xspace}
\usepackage{multirow}
\usepackage{multicol}
\usepackage{fancyhdr}

\newcommand{\name}{CLIP-UP\xspace}

\title{\name: A Simple and Efficient Mixture-of-Experts CLIP Training Recipe with Sparse Upcycling}
\author{
  Xinze Wang \hspace{0.3em}
  Chen Chen  \hspace{0.3em}
  Yinfei Yang  \hspace{0.3em}
  Hong-You Chen \hspace{0.3em}
  Bowen Zhang \vspace{0.1em}\\
  \bf{Aditya Pal} \hspace{0.3em}
  \bf{Xiangxin Zhu} \hspace{0.3em}
  \bf{Xianzhi Du} \vspace{0.5em} \\
  Apple \vspace{0.1em} \\
  \texttt{xinze\_wang@apple.com}
}

\begin{document}
\maketitle

\begin{abstract}
Mixture-of-Experts (MoE) models are crucial for scaling model capacity while controlling inference costs. While integrating MoE into multimodal models like CLIP improves performance, training these models is notoriously challenging and expensive. We propose \textbf{CLIP-Up}cycling (\textbf{\name}), an efficient alternative training strategy that converts a pre-trained dense CLIP model into a sparse MoE architecture. Through extensive experimentation with various settings and auxiliary losses, we demonstrate that \name significantly reduces training complexity and cost. Remarkably, our sparse CLIP B/16 model, trained with \name, outperforms its dense counterpart by 7.2\% and 6.6\% on COCO and Flickr30k text-to-image Recall@1 benchmarks respectively. It even surpasses the larger CLIP L/14 model on this task while using only 30\% of the inference FLOPs. We further demonstrate the generalizability of our training recipe across different scales, establishing sparse upcycling as a practical and scalable approach for building efficient, high-performance CLIP models.
\end{abstract}

\section{Introduction}

CLIP~\cite{radford2021learningtransferablevisualmodels,jia2021scaling} has become foundational across domains such as image classification, multimodal retrieval, and AI-driven multimodality content generation~\cite{Zhou_2022,rao2022densecliplanguageguideddenseprediction,gan2022visionlanguagepretrainingbasicsrecent,ramesh2021zeroshottexttoimagegeneration,liu2023visualinstructiontuning}. As applications grow, scaling CLIP becomes essential. Most efforts focus on enlarging dense models~\cite{Cherti_2023}, which improves performance but incurs high computational and inference costs.

\begin{figure}[t!]
\begin{center}
\centerline{\includegraphics[width=1.0\columnwidth]{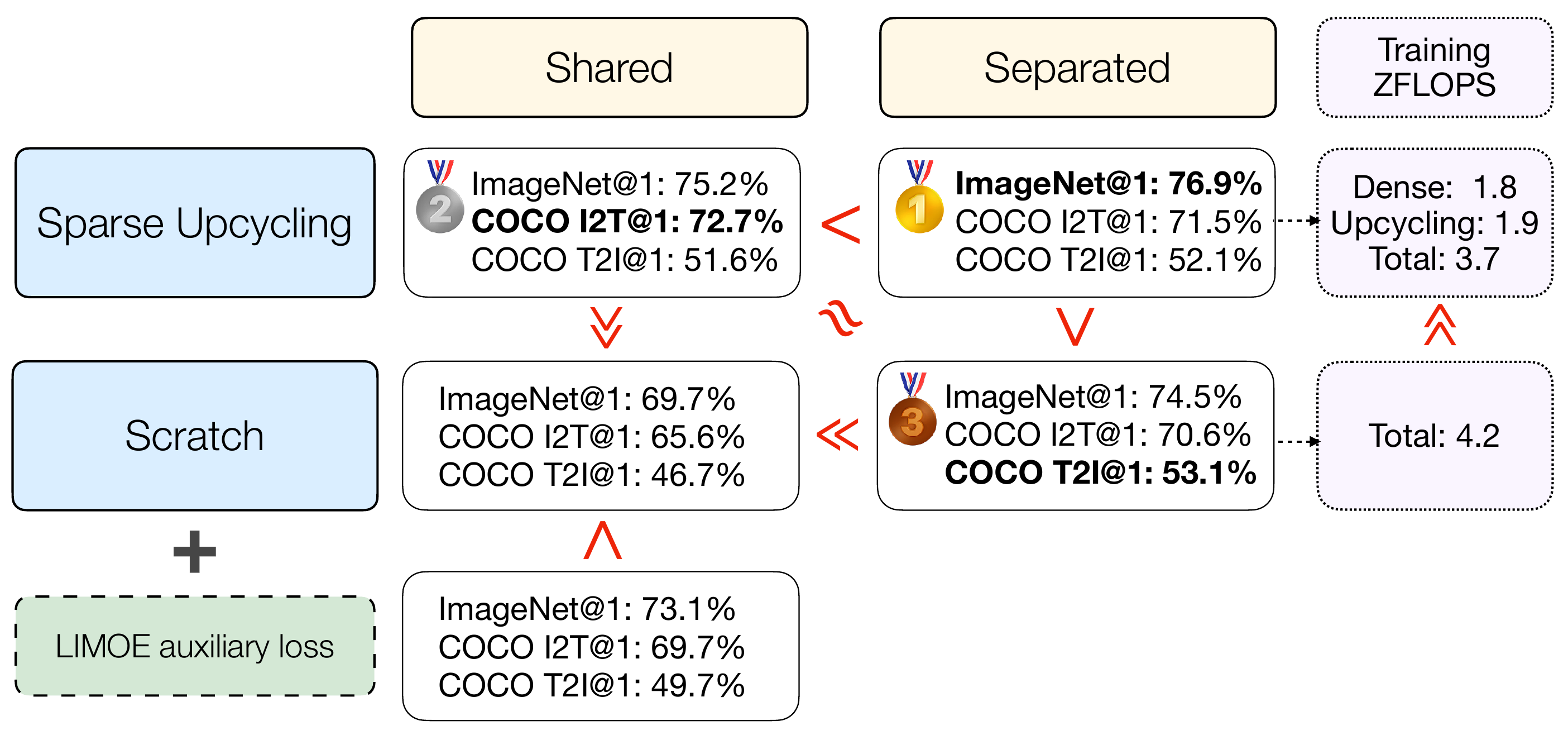}}
\caption{\textbf{Our proposed MoE CLIP pre-training recipe.} We highlight key factors for efficient training, including backbone sharing, training from scratch vs. sparse upcycling, and auxiliary losses.
A detailed analysis is provided in Section~\ref{comparison-methodology} and Section~\ref{limoe-loss}. 
}
\label{fig:strategy-compare}
\end{center}
\vskip -0.3in
\end{figure}


An efficient alternative is sparse modeling with Mixture-of-Experts (MoE)~\cite{mustafa2022multimodalcontrastivelearninglimoe, shazeer2017outrageouslylargeneuralnetworks}.
However, training MoE-based CLIP models like LIMOE~\cite{mustafa2022multimodalcontrastivelearninglimoe} from scratch remains expensive and often requires auxiliary losses for stability. For instance, LIMOE outperforms dense CLIP but demands 1.35× more training FLOPs~\cite{mustafa2022multimodalcontrastivelearninglimoe}.

To address this, we explore sparse upcycling~\cite{komatsuzaki2023sparseupcyclingtrainingmixtureofexperts}, which initializes MoE layers from a pre-trained dense model. As shown in Figure~\ref{fig:strategy-compare}, our extensive experiments demonstrate that sparse upcycling with a separated backbone achieves the best performance while reducing training ZFLOPs from 4.2 to 3.7 compared to training from scratch. Although LIMOE’s entropy losses~\cite{mustafa2022multimodalcontrastivelearninglimoe} improve shared-backbone models trained from scratch, they still underperform other setups. Section~\ref{comparison-methodology} details these strategies and the effects of auxiliary losses.



In contrast, we propose \textbf{\name}, a single-stage sparse upcycling method for CLIP. By leveraging pre-trained weights, \name provides a warm start that boosts efficiency and surpasses both dense continued training and sparse-from-scratch methods across model scales.\footnote{Concurrent work CLIP-MoE~\cite{zhang2024clipmoebuildingmixtureexperts} also explores MoE upcycling, using cluster-and-contrast learning to initialize experts. However, it requires additional training stages per expert, making it difficult to scale. }

  
  

Our main contributions are:
\begin{enumerate}[nosep,topsep=0pt,parsep=0pt,partopsep=0pt, leftmargin=*]
\item We introduce \name, a simple and effective training recipe for MoE CLIP models via sparse upcycling, avoiding complex auxiliary losses and outperforming existing methods across shared and separated backbones.
\item \name significantly improves performance on text-image retrieval, surpassing dense CLIP by 7.2\% and 5.5\% (recall@1) on COCO and Flickr30K, respectively, with a B/16 backbone.
\item We demonstrate \name's scalability from B/32 to L/14 and provide insights into key factors and challenges to inform future design.
\end{enumerate}

\section{\name}
\label{method}

\begin{figure}[t!]
\vskip -0.1in
\begin{center}
\centerline{\includegraphics[width=0.48\textwidth]{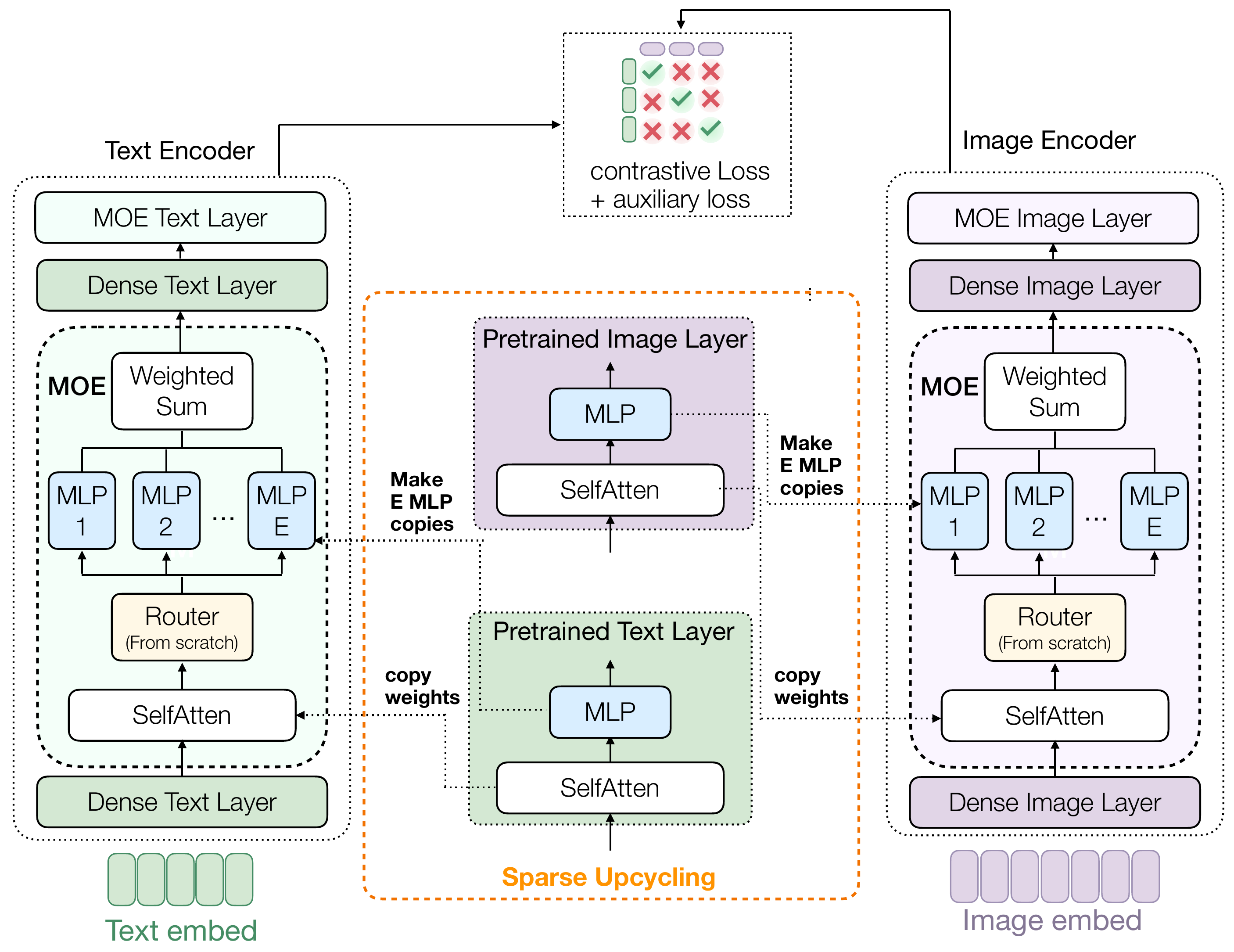}}
\caption{\name overview with sparse upcycling initialization. Selected MLP layers are replaced with MoE layers, initialized from the dense checkpoint, while routers are randomly initialized.
}
\label{fig:clip-moe-flow}
\end{center}
\vskip -0.3in
\end{figure}

Figure~\ref{fig:clip-moe-flow} illustrates the \name architecture and training strategy. We detail both in this section.


\subsection{CLIP}
Given $n$ pairs of image and text captions $\{(\mathbf{I}_j, \mathbf{T}_j)\}_{j=1}^{n}$, CLIP~\cite{radford2021learningtransferablevisualmodels} learns image and text embeddings $f(\mathbf{I}_j)$ and $g(\mathbf{T}_j)$ using a contrastive loss. 
With batch size $B$ and temperature $\theta$, the loss is
\begin{equation}
\begin{aligned}
\mathcal{L}_\text{Contrastive} &= -\frac{1}{2B}\sum_{j=1}^B (\log \frac{e^{\text{sim}(f(\mathbf{I}_j), g(\mathbf{T}_j)) / \theta}}{\sum_{k=1}^{B} e^{\text{sim}(f(\mathbf{I}_j), g(\mathbf{T}_k)) / \theta}} \\ 
&+ \log \frac{e^{\text{sim}(f(\mathbf{I}_j), g(\mathbf{T}_j)) / \theta}}{\sum_{k=1}^{B} e^{\text{sim}(f(\mathbf{I}_k), g(\mathbf{T}_j)) / \theta}})
\end{aligned}
\tag{1}
\end{equation}




\subsection{CLIP with Mixture-of-Experts upcycling} 

Each MoE layer consists of $E$ MLP experts and a router that activates the top-$K$ experts per input token based on predicted gating logits. Let $\mathbf{X}_j \in \mathbb{R}^{D}$ be the input for the $j$-th token, $\mathbf{G}_{e,j} \in \mathbb{R}^{D}$ the gating logits, and $\mathbf{W}_e \in \mathbb{R}^{D}$ the router weights for expert $e$. The output $\text{MoE}(\mathbf{X}_j)$ is computed as:
\[
\text{MoE}(\mathbf{X}_j) = \mathbf{X}_j + \sum_{e \in \text{Top-K}}G_{e,j}\text{MLP}_e(\mathbf{X}_j)
\]
\[
\hspace{0.5 cm} G_{e,j} = 
\begin{cases}
    \text{Softmax}(\mathbf{W}_e^T \mathbf{X}_j), &\text{if}\ e \in \text{Top-K}, \tag{2} \\
    0, & \text{otherwise},
\end{cases}
\]

Each expert is assigned a fixed buffer capacity~\cite{fedus2022switchtransformersscalingtrillion}, allowing it to process a limited number of tokens at a time. With capacity factor $C$, batch tokens $B_t$, the capacity per expert is $B_e = (B_t / E) \times C$. This ensures computational efficiency and effective resource management. 
Tokens are assigned to experts on a "first-come-first-serve" basis~\cite{fedus2022switchtransformersscalingtrillion}. This simple mechanism avoids prioritization overhead while efficiently distributing tokens across experts.



\textbf{Auxiliary loss.}
Simplified token assignment reduces overhead but risks imbalanced token distribution, leading to token dropping and performance degradation~\cite{zeng2024turnwasteworthrectifying}. To mitigate this, we adopt an auxiliary loss~\cite{zoph2022stmoedesigningstabletransferable} combining load balance loss and router $z$-loss with scaling factors $\alpha$ and $\beta$. The load balance loss promotes uniform token allocation across experts. For a sequence of length $S$, it is defined as:


\[
\mathcal{L}_\text{Balance} = \alpha \cdot \sum_{e=1}^E R_e \cdot P_e \tag{3}
\]

where $R_e = \frac{E}{K \cdot S} \sum_{j=1}^S \mathbbm{1}(\text{Token $j$} \rightarrow \text{Expert $e$})$
and 
$ P_e = \frac{1}{S} \sum_{j=1}^S \mathbf{G}_{e,j} $
, denoting the token assignment ratio and average router probability for expert $e$ respectively.

The router $z$-loss stabilizes gating by regularizing router logits to keep outputs within a reasonable range. It is defined as:
\[
\mathcal{L}_\text{Router} = \beta \cdot \frac{1}{S} \sum_{j=1}^{S} \left( \log \sum_{e=1}^{E} e^{\mathbf{G}_{e,j}} \right)^2
\tag{4}
\]


\textbf{LIMOE auxiliary loss.} 
We experimented with LIMOE’s local and global entropy losses~\cite{mustafa2022multimodalcontrastivelearninglimoe}, tuning hyperparameters accordingly. While LIMOE auxiliary loss improves shared backbone trained from scratch, it underperforms in other settings. Therefore, we use load balance and router-$z$ losses as our auxiliary loss.

\subsection{Sparse Upcycling Training}
Sparse upcycling begins with a pre-trained dense CLIP, replacing selected MLP layers with MoE layers—experts initialized from the dense weights and routers randomly initialized. All other layers remain unchanged. The model is then fine-tuned with slightly reduced learning rate and weight decay for improved stability, as shown in Figure~\ref{fig:clip-moe-flow}.

\section{Experiments}
\label{sec:exp}

\paragraph{Datasets.} We trained both the initial dense CLIP checkpoint, \name, and the baseline model on the same paired image-text datasets—WIT-300M~\citep{wu2024mofilearningimagerepresentations} and DFN-5B~\citep{fang2023datafilteringnetworks}. Evaluation was performed on ImageNet~\citep{deng2009imagenet,pmlr-v119-shankar20c} for classification and on COCO~\citep{lin2014microsoft} and Flickr30K~\citep{flickr30k} for image-text retrieval, with additional benchmarks provided in the Appendix~\ref{appendix-imagenet-variants}. The input image resolution is 224 for all of the datasets. 

\input{tabels/methodology-compare-no-limoe}

\input{tabels/zero_shot_metrics_flops}
\paragraph{Setup.}
We train a dense CLIP model for 440k steps, then upcycle it into an MoE version with 350k additional steps. Both use AdamW with a 32k batch size; the dense model uses a learning rate of $5 \times 10^{-4}$ and weight decay of 0.2, reduced to $5 \times 10^{-5}$ and 0.05 for upcycling. In the MoE model, half of the Transformer’s MLP layers follow an alternating [dense, sparse] pattern~\cite{zoph2022stmoedesigningstabletransferable, du2022glamefficientscalinglanguage}, each sparse layer using 8 experts with top-2 routing. Router loss coefficients $\alpha = 0.01$ and $\beta = 0.001$ balance expert usage without dominating training~\cite{zoph2022stmoedesigningstabletransferable, xue2024openmoeearlyeffortopen}. For fair comparison, we also train a dense CLIP for 790k steps using the same settings.


\subsection{Recipe Study}
\label{comparison-methodology}
We compare shared vs. separated backbones and training from scratch vs. sparse upcycling using the CLIP-B/16 model, with the shared setup using 16 experts to match the separated configuration (8 per modality). As shown in Table \ref{tab:methodology-compare-no-limoe}, the separated backbone with sparse upcycling delivers the best overall performance due to dedicated parameters per modality, while the shared backbone sees greater relative gains from sparse upcycling. Overall, sparse upcycling consistently outperforms training from scratch, demonstrating CLIP-UP’s versatility and efficiency across configurations.

\subsection{Impact of LIMOE auxiliary loss.} 
\label{limoe-loss}

\begin{figure}[t!]  
    \centering  
    \includegraphics[width=1.0\linewidth]{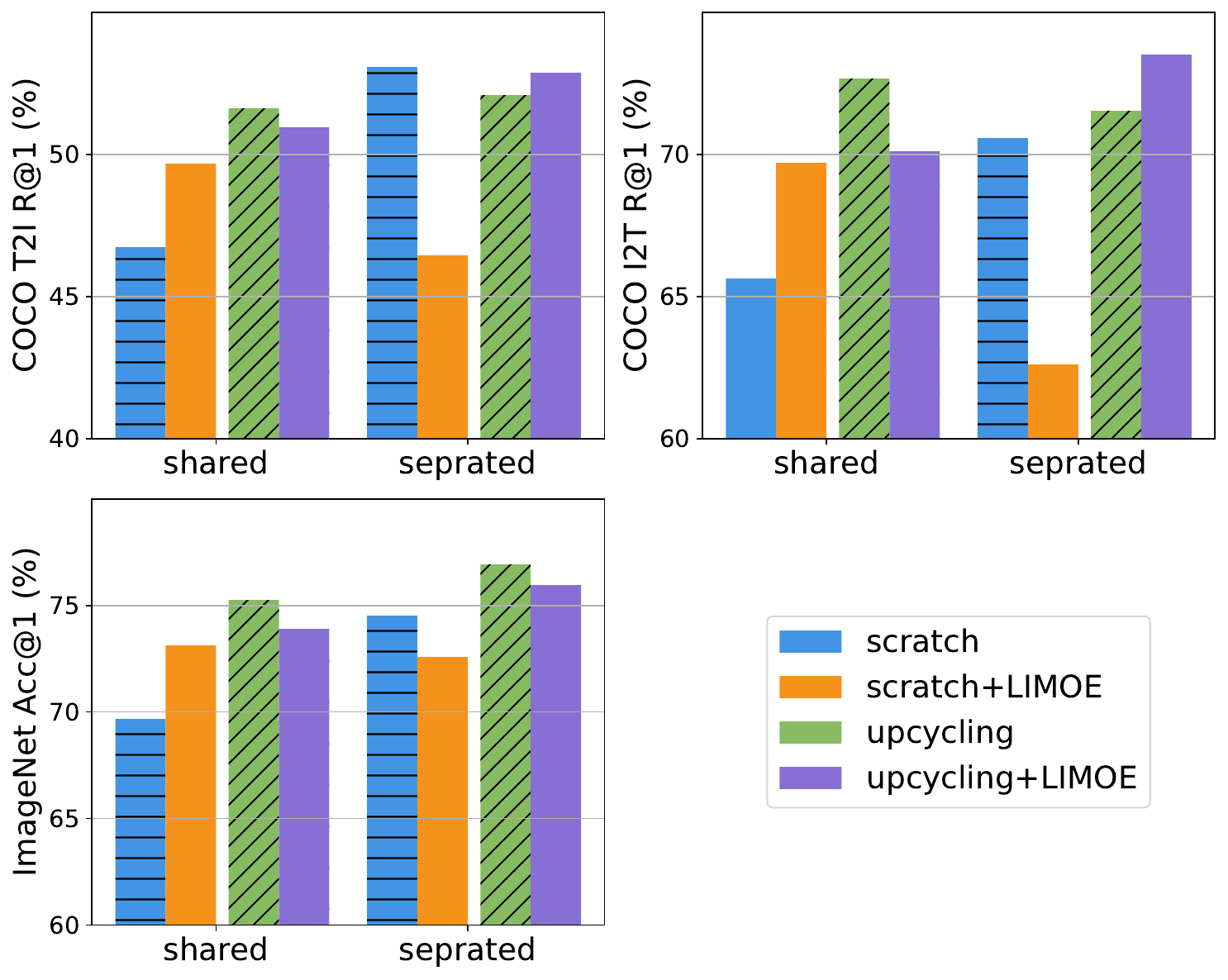}
    \vspace{-10pt}
    \caption{
    Impact of LIMOE auxiliary loss under different training setups. Adding LIMOE loss sometimes causes instability, especially with unshared backbones, while our upcycling recipe remains more robust.
    } 
    \label{fig:limoe-compare}  
\end{figure}

We examine the LIMOE auxiliary loss by setting $\tau = 6$ in the global entropy loss~\cite{mustafa2022multimodalcontrastivelearninglimoe}, encouraging use of at least six experts per modality, and tuning the loss weight for our setup. As shown in Figure \ref{fig:limoe-compare}, it improves ImageNet and COCO performance with a shared backbone trained from scratch, consistent with prior work ~\cite{mustafa2022multimodalcontrastivelearninglimoe}, but still underperforms compared to other configurations without it. Applying the loss to our best setup (Separated-Upcycle) slightly boosts text-image retrieval but falls short on ImageNet zero-shot classification.

These auxiliary losses also increase training complexity due to more hyperparameters. As LIMOE loss didn’t work reliably across all setups, we use load balance and router-$z$ losses to simplify tuning under resource constraints.

\subsection{Final Model Evaluation and Baselines}
Based on previous results, we adopt the separated backbone with sparse upcycling as the default setup and evaluate \name across model sizes from B/32 to L/14. Table \ref{zero-shot-metrics-flops-steps} compares zero-shot classification and retrieval performance against dense CLIP models trained for the same number of steps. While extending dense CLIP training from 440k to 790k steps yields minor gains, \name shows consistent, significant improvements across scales, especially in retrieval. Notably, \name B/32 uses only 47\% of the inference GFLOPS yet outperforms dense CLIP B/16 in COCO T2I recall@1 by 2.4\%, while \name B/16 uses just 31\% of the GFLOPS and surpasses dense CLIP L/14 by 1.9\%. These results demonstrate the efficiency and effectiveness of sparse upcycling for scaling CLIP models.

\begin{figure}[ht]  
    \centering  
    \includegraphics[width=0.95\linewidth]{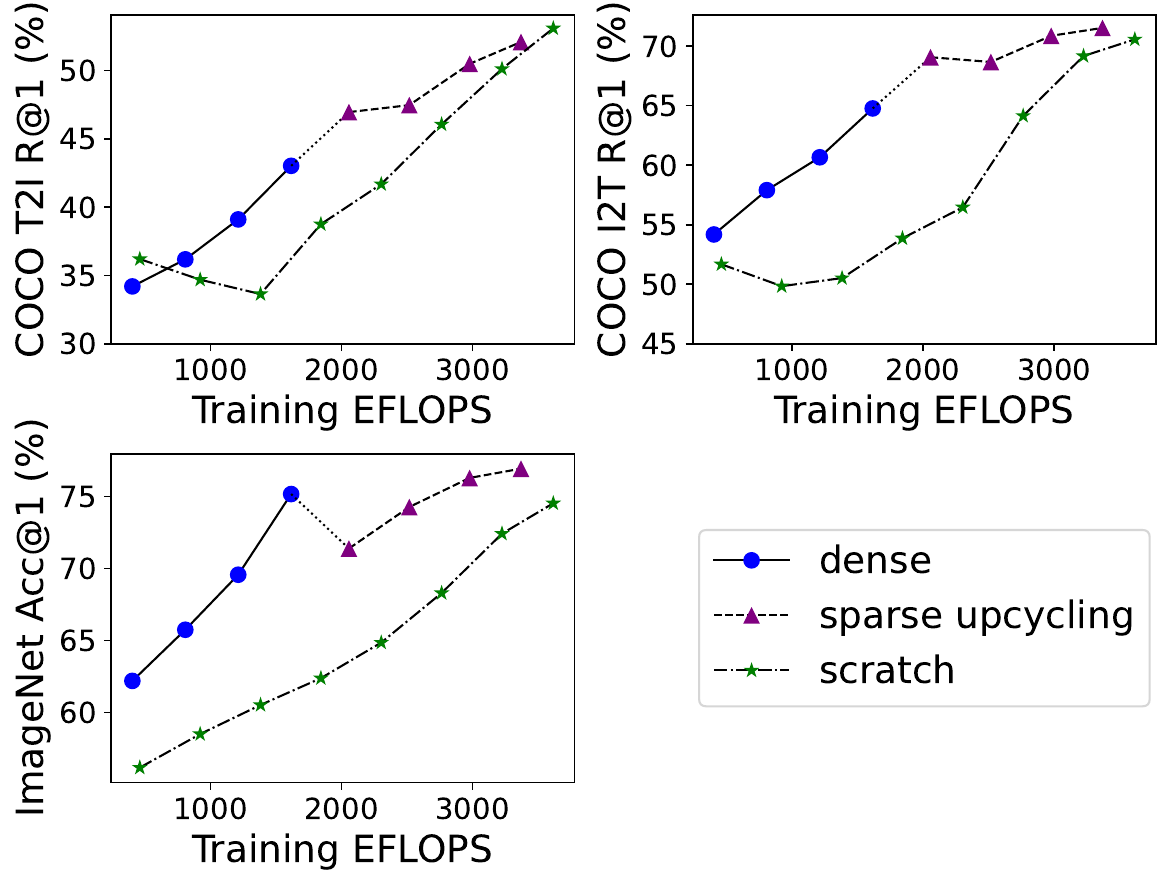}
    \caption{Performance vs. training EFLOPS for CLIP-UP and sparse-from-scratch model on CLIP B/16.
    }
    \label{fig:scratch-comparison}  
\vskip -0.15in
\end{figure} 

\subsection{Training Efficiency}

To highlight the effectiveness of upcycle training, Figure \ref{fig:scratch-comparison} compares dense pretraining + upcycling with training from scratch. The pretrained dense model provides a strong starting point, while training from scratch requires significantly more compute to match \name's performance—especially on COCO image-to-text and ImageNet. Although sparse upcycling initially causes a performance drop on ImageNet due to reconfiguration, \name consistently outperforms the scratch baseline, demonstrating better efficiency and overall performance.

\section{Conclusions}
\label{conclusion}
We present \name, an efficient CLIP training strategy that combines MoE with sparse upcycling. Extensive experiments show it reduces training costs and inference FLOPs while improving performance across scales, even outperforming larger dense models. Ablation studies shown in Appendix~\ref{appendix-ablation-study} further validate key design choices, highlighting \name's practicality and scalability.
\newpage
\section{Limitation}
While our proposed method demonstrates strong performance improvements in retrieval tasks such as COCO and Flickr30K, it reveals a trade-off with classification performance on ImageNet and its variants. Specifically, we have not yet identified a training configuration that yields significant gains across both retrieval and classification simultaneously. Our current best setup prioritizes retrieval effectiveness, achieving notable improvements on COCO and Flickr30K, but leads to only marginal gains on ImageNet.

We discuss this trade-off in more detail in Appendix B.3, highlighting the role of the expert capacity factor in shaping task-specific performance. In particular, we provide examples showing how ImageNet and COCO respond differently to token dropping under varying expert capacities, which we believe contributes to the observed trade-off. While these insights offer a preliminary understanding, we are still exploring more effective strategies to better balance retrieval and classification performance.

\section{Acknowledgement}
We thank Wentao Wu, Haotian Zhang, and many others for their invaluable help and feedback. 


\input{acl_latex.bbl}

\appendix
\newpage
\onecolumn


\section{Training Details}
Below, we provide detailed training hyper-parameters and setups for dense CLIP (weights are used for sparse upcycling), sparse CLIP trained from scratch, and \name.

\subsection{Training hyper-parameters}
We primarily follow \cite{radford2021learningtransferablevisualmodels} for hyper-parameter selection, using the WIT-3000M~\cite{wu2024mofilearningimagerepresentations} and DFN-5B~\cite{fang2023datafilteringnetworks} training datasets. Table \ref{tab:hyper-parameter} summarizes the hyper-parameters for all experiments, including MoE-specific configurations and parameters for dense CLIP, sparse CLIP, and \name.

\input{tabels/appendix_hyper_parameters}

\newpage
\section{Ablation study}
\label{appendix-ablation-study}

\subsection{MoE added to single or multiple modalities} 
\begin{figure*}[ht]  
    \centering  
    \vskip -0.1in
    \includegraphics[width=0.95\linewidth]{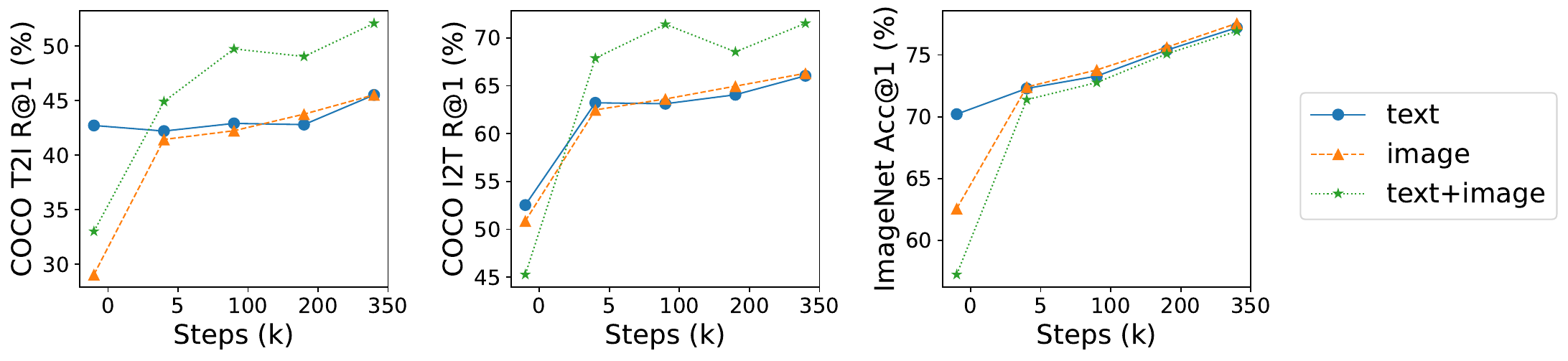}
    \caption{\name with MoE upcycling for only the text encoder, image encoder, or both. We observe upcycling both the image and text encoders into MoE generally helps, especially for retrieval tasks.} 
    \label{fig:moe-modality}  
    \vskip -0.1in
\end{figure*}

In the \name model, the MoE setup is applied to both the text encoder and image encoder. We also explore the effect of adding MoE layers to only one modality while keeping the other modality fully dense. The COCO retrieval and ImageNet results for these configurations are shown in Figure \ref{fig:moe-modality}, measured across different training steps.

We observe that the initial performance of newly upcycled models tends to decline compared to the starting dense model, regardless of where the MoE setup is applied. However, all configurations recover after approximately 5k training steps. Applying MoE layers to both modalities leads to a more significant initial drop on average. When MoE layers are applied to only one modality (either image or text), the final performance remains comparable. Notably, the initial performance drop is more pronounced when MoE layers are applied to the image modality, suggesting that the model is more sensitive to changes in image representations.

\subsection{Expert capacity factor}
\begin{figure}[h]  
    \centering  
    \includegraphics[width=1.0\linewidth]{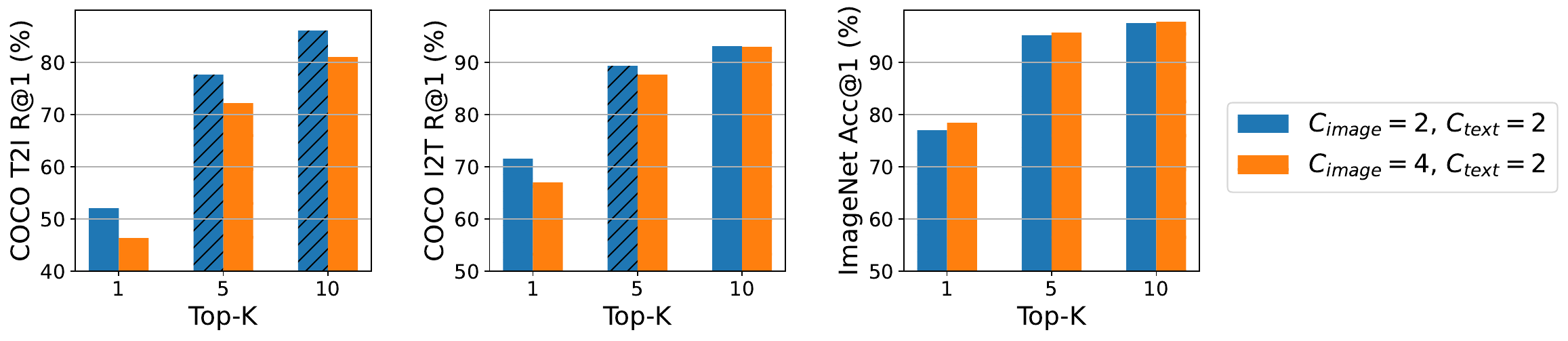}
    \vskip -0.1in
    \caption{Effects of different expert capacity $C$.} 
    \label{fig:capacity-compare}  
\end{figure}

Intuitively, the number of tokens processed by each expert plays a crucial role in determining model quality. In \name, this is controlled by the expert capacity factor, denoted as $C$. A higher $C$ results in less token dropping, thereby reducing the initial quality drop. However, this doesn't necessarily guarantee a higher final model quality. By default, we set the capacity factor to 2.0 for both modalities. As shown in Figure \ref{fig:capacity-compare}, increasing $C_{image}$ to 4.0 significantly boosts the ImageNet zero-shot metrics. However, this adjustment results in a noticeable drop in performance on COCO and Flickr30k retrieval tasks.

\label{expert_capacity}
\begin{figure}[ht]  
    \centering  
    \begin{subfigure}
        \centering
        \includegraphics[width=0.8\linewidth]{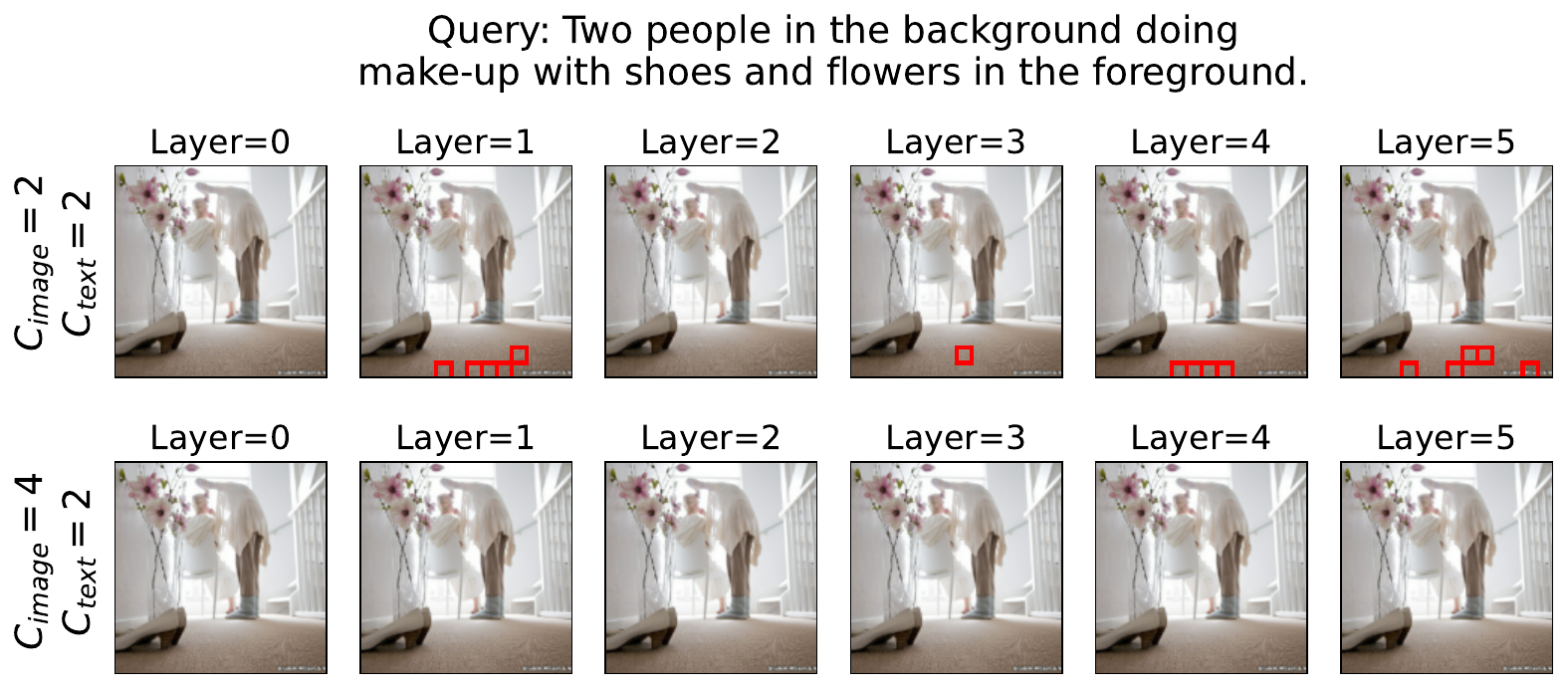}
    \end{subfigure}
    \begin{subfigure}
        \centering
        \includegraphics[width=0.8\linewidth]{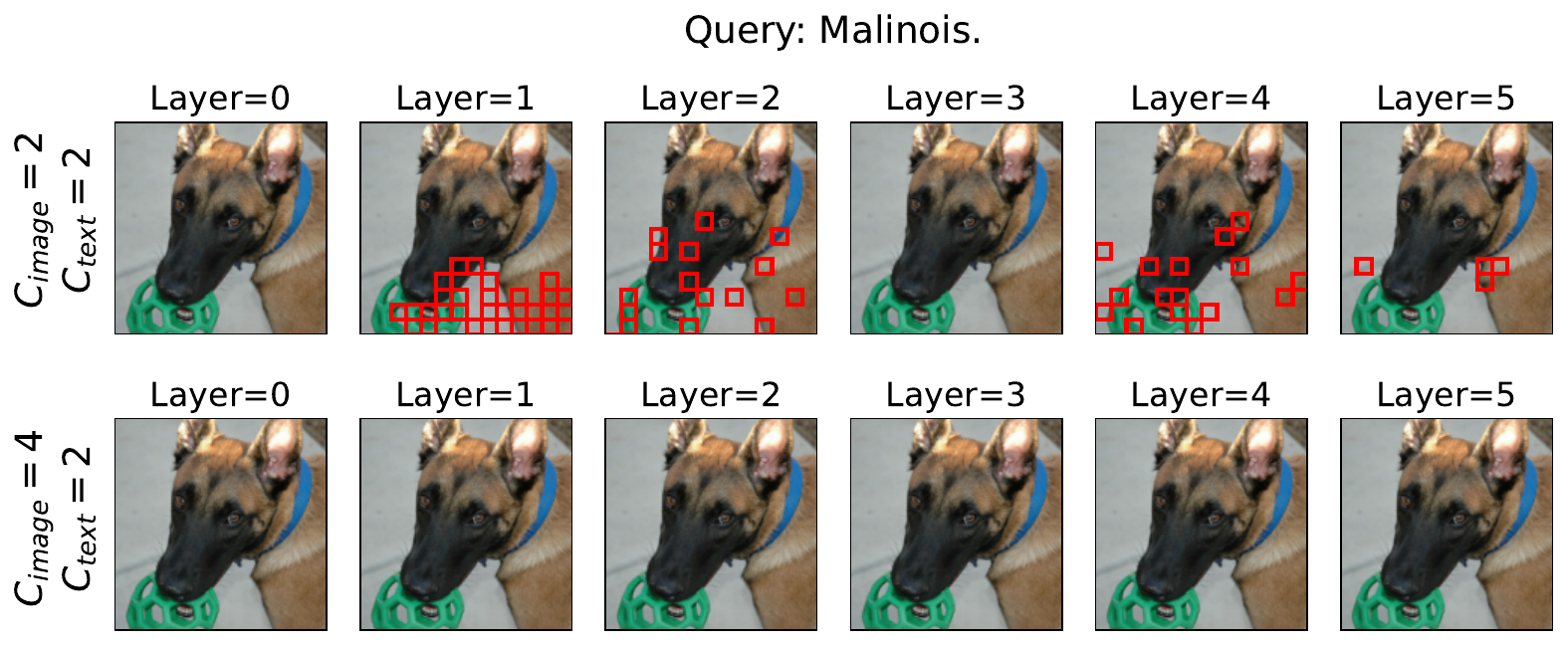}
    \end{subfigure}
    \vskip -0.1in
    \caption{Visualization of image tokens dropped by the router (i.e., not being assigned to any expert due to capacity constraint) on COCO (Top) and ImageNet (Bottom).}
    \label{fig:imagenet-dropped-image-token}  
\end{figure}

Figure \ref{fig:imagenet-dropped-image-token} visualizes the behavior of dropped image tokens under different capacity settings of COCO and Imagenet. The red squares represent the dropped image tokens. With a higher $C_{image}$, fewer image tokens are discarded. This benefits ImageNet performance since the dataset primarily consists of single-object images. Retaining more tokens allows the model to focus on the key features. In contrast, COCO images often depict complex scenes with multiple objects, but not all of which are relevant to the paired captions. Dropping less important image tokens helps the model concentrate on the most important objects, which explains the drop in COCO performance when fewer tokens are discarded.
We leave the further study of capacity factor for different tasks to a future work.

\subsection{Normalize gating weights before or after routing.} 
To mitigate the initial quality drop observed when applying sparse upcycling, we experimented with normalizing the router output logits after routing. This ensures the remaining gating weights are normalized to sum to 1, even when some tokens are dropped due to expert capacity constraints. The intuition behind this approach is that in the dense model, each token was previously processed by a single expert MLP.

\begin{figure}[ht]  
    \centering  
    \includegraphics[width=1.0\linewidth]{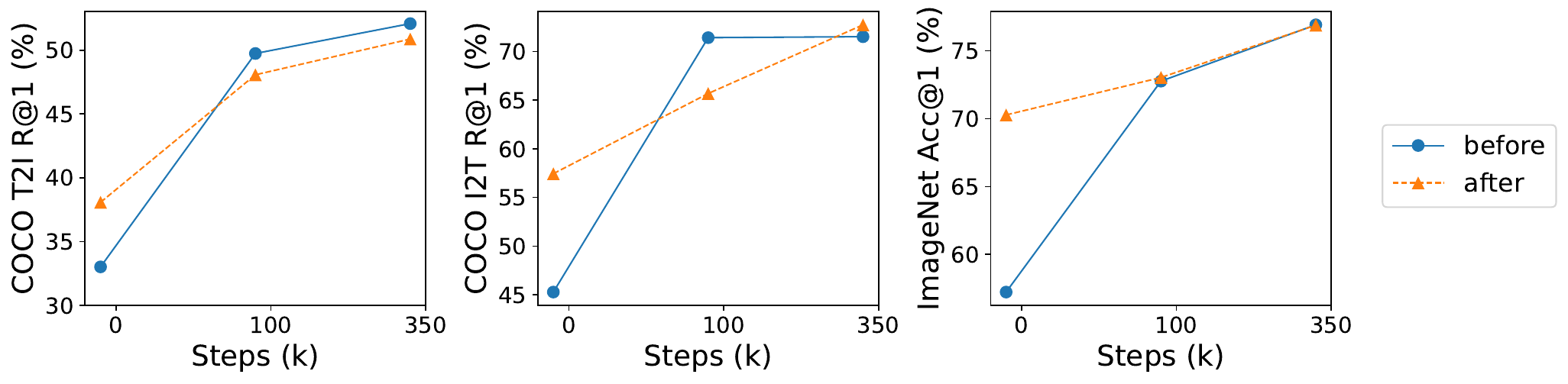}
    \caption{Model performance for gating normalization applied before or after routing} 
    \label{fig:normalization-compare}  
\end{figure}

As shown in Figure \ref{fig:normalization-compare}, normalizing gating weights post-routing helps reduce the initial quality drop. However, in terms of final model performance, this approach shows improved results in image-to-text retrieval, but performs worse in text-to-image retrieval. 

A possible explanation for this discrepancy is that post-routing normalization maintains the magnitude of all remaining tokens, which benefits the image encoder, as most image tokens are informative. In contrast, text encoder often deals with padding tokens, and reducing the magnitude of these tokens can enhance the text encoder's ability to focus on meaningful content. It also aligns with the finding that the initial quality drop is the biggest when adding MoE layers into image modality only.

\newpage
\section{Tabular results}
\subsection{Comparison of model architectures and impact of LIMOE auxiliary loss} 

All results for different model architectures, with and without LIMOE auxiliary loss, as discussed in Section \ref{comparison-methodology}.

\input{tabels/appendix_full_method_compare_with_LIMOE}

\subsection{Performance on ImageNet variants} 
\label{appendix-imagenet-variants}
To complement the performance comparison in Table~\ref{zero-shot-metrics-flops-steps}, we additionally evaluated our models on several ImageNet variants, including ImageNet-V2~\cite{recht2019imagenetclassifiersgeneralizeimagenet}, ImageNet-A~\cite{hendrycks2021naturaladversarialexamples}, and ImageNet-R~\cite{hendrycks2021facesrobustnesscriticalanalysis}. As shown in Table~\ref{appendix_imagenet_variants}, the performance trends on these datasets are consistent with those observed on the original ImageNet benchmark.
\input{tabels/appendix_imagenet_variants}

\subsection{Comparison of MoE added to single modality or both modalities} 

All results from Figure \ref{fig:moe-modality} to compare MoE added into different modalities. 

\input{tabels/appendix_modality_moe}

\subsection{Comparison of capacity factor} 

As discussed in Section~\ref{expert_capacity}, expert capacity factor $C$ plays a crucial role in balancing classification performance on ImageNet and its variants with retrieval performance on Flickr and COCO. As shown in Table~\ref{tab:appendix-capacity-compare-imagenet-variants}, increasing $C$ from 2.0 to 4.0 consistently improves accuracy on ImageNet and its variants, but at the expense of COCO performance—highlighting an inherent trade-off. 

\input{tabels/appendix_capacity_factor_imagenet_variants}

\newpage
\section{Router Analysis}

\subsection{Routing distribution}
\begin{figure}[ht]  
    \centering  
    \begin{subfigure}
        \centering
        \includegraphics[width=0.8\linewidth]{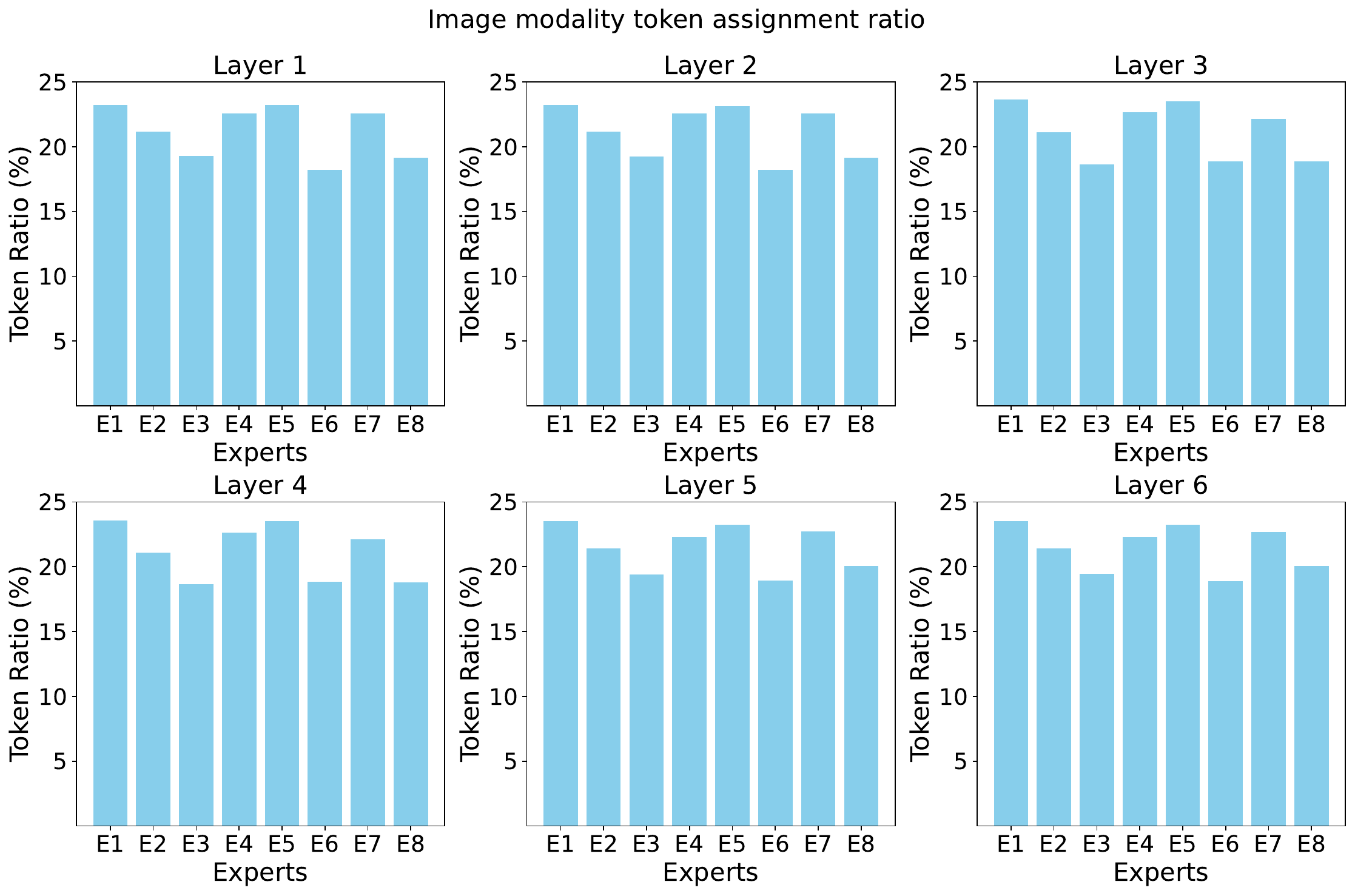}
    \end{subfigure}
    \begin{subfigure}
        \centering
        \includegraphics[width=0.8\linewidth]{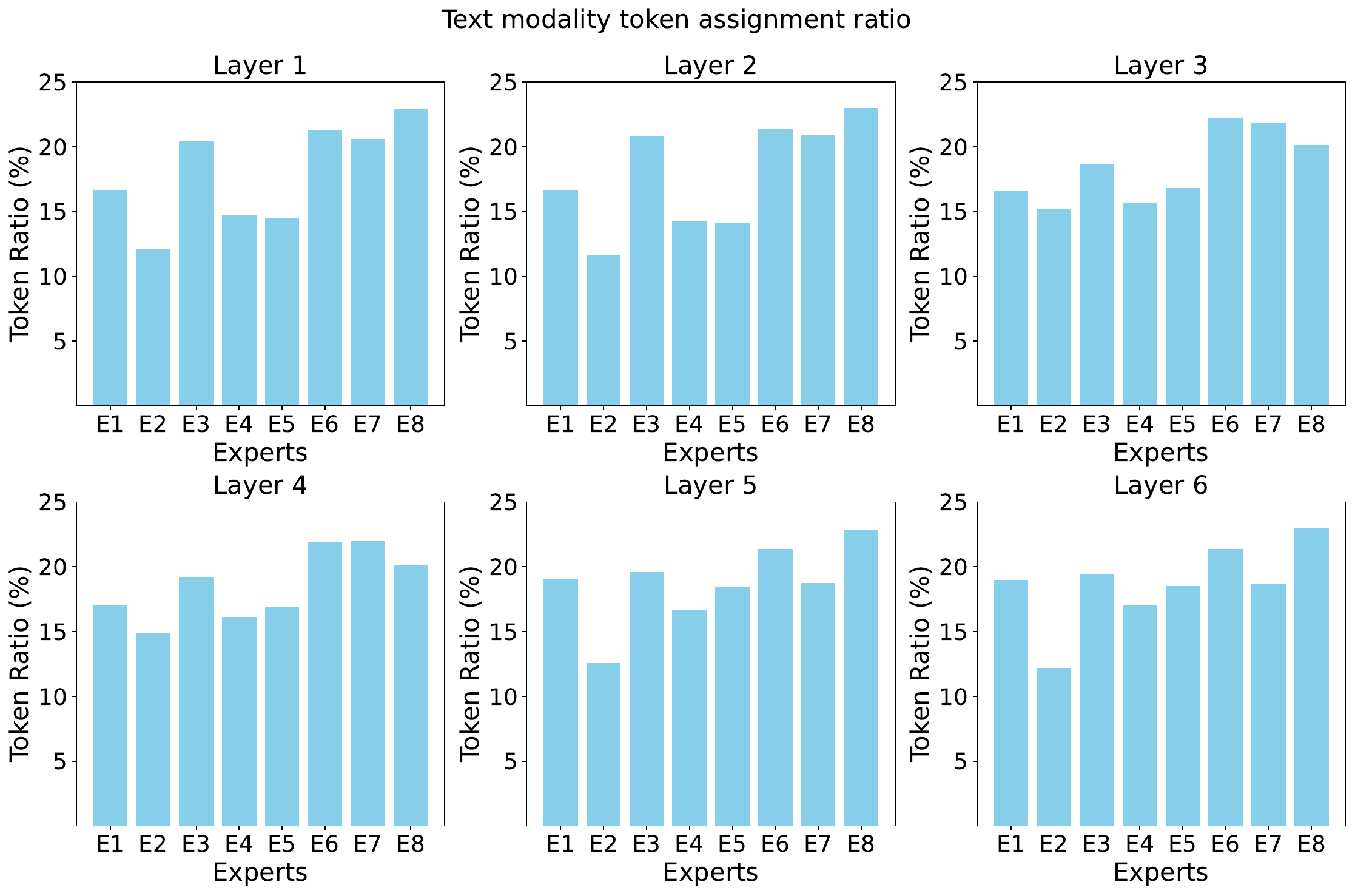}
    \end{subfigure}
    \vskip -0.1in
    \caption{Visualization of the token assignment ratios to each expert in each layer by the router on ImageNet — image tokens (top) and text tokens (bottom).}
    \label{fig:image-text-assign-ratio}  
\end{figure}

\begin{figure}[ht]  
    \centering  
    \begin{subfigure}
        \centering
        \includegraphics[width=0.8\linewidth]{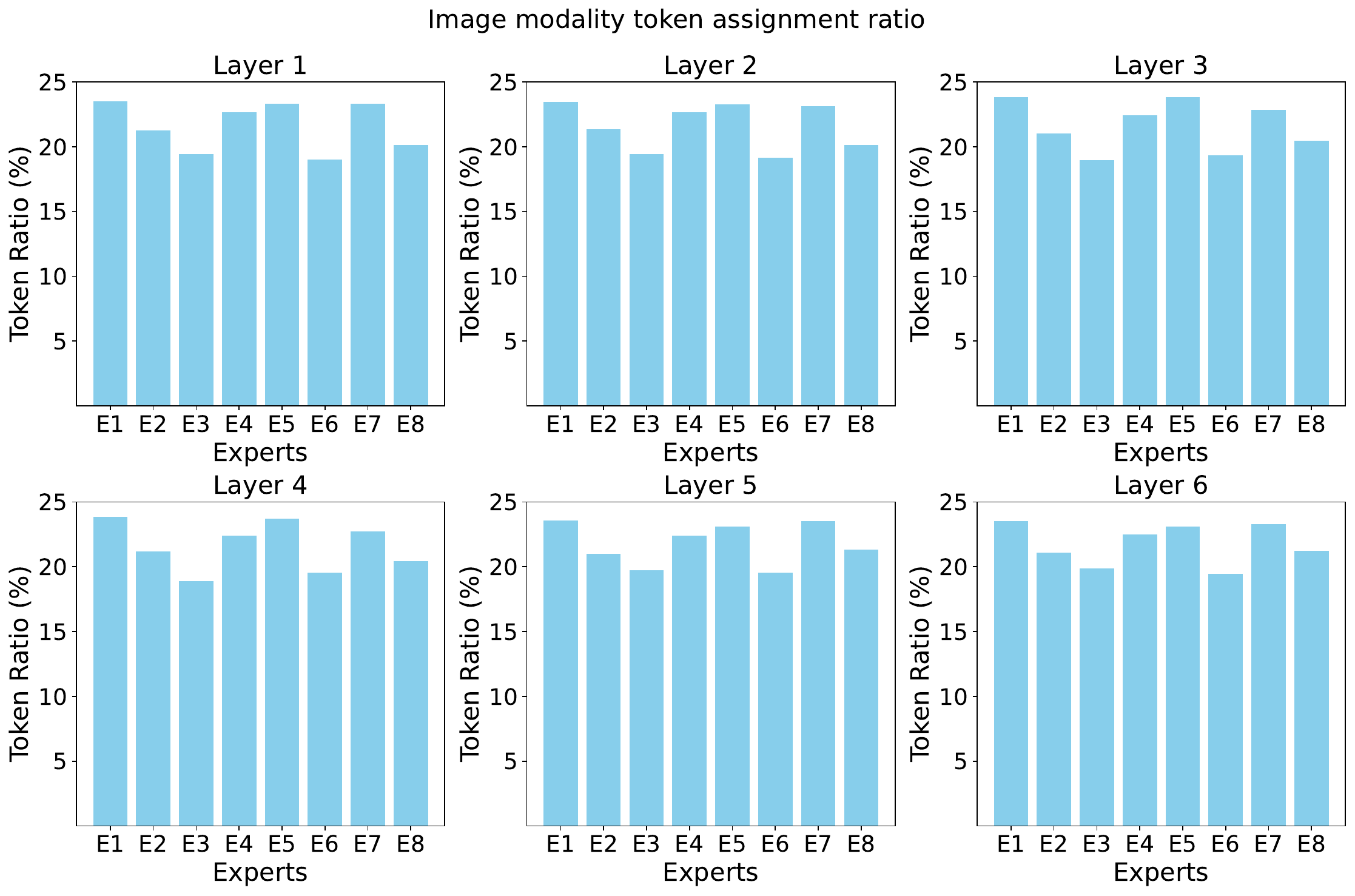}
    \end{subfigure}
    \begin{subfigure}
        \centering
        \includegraphics[width=0.8\linewidth]{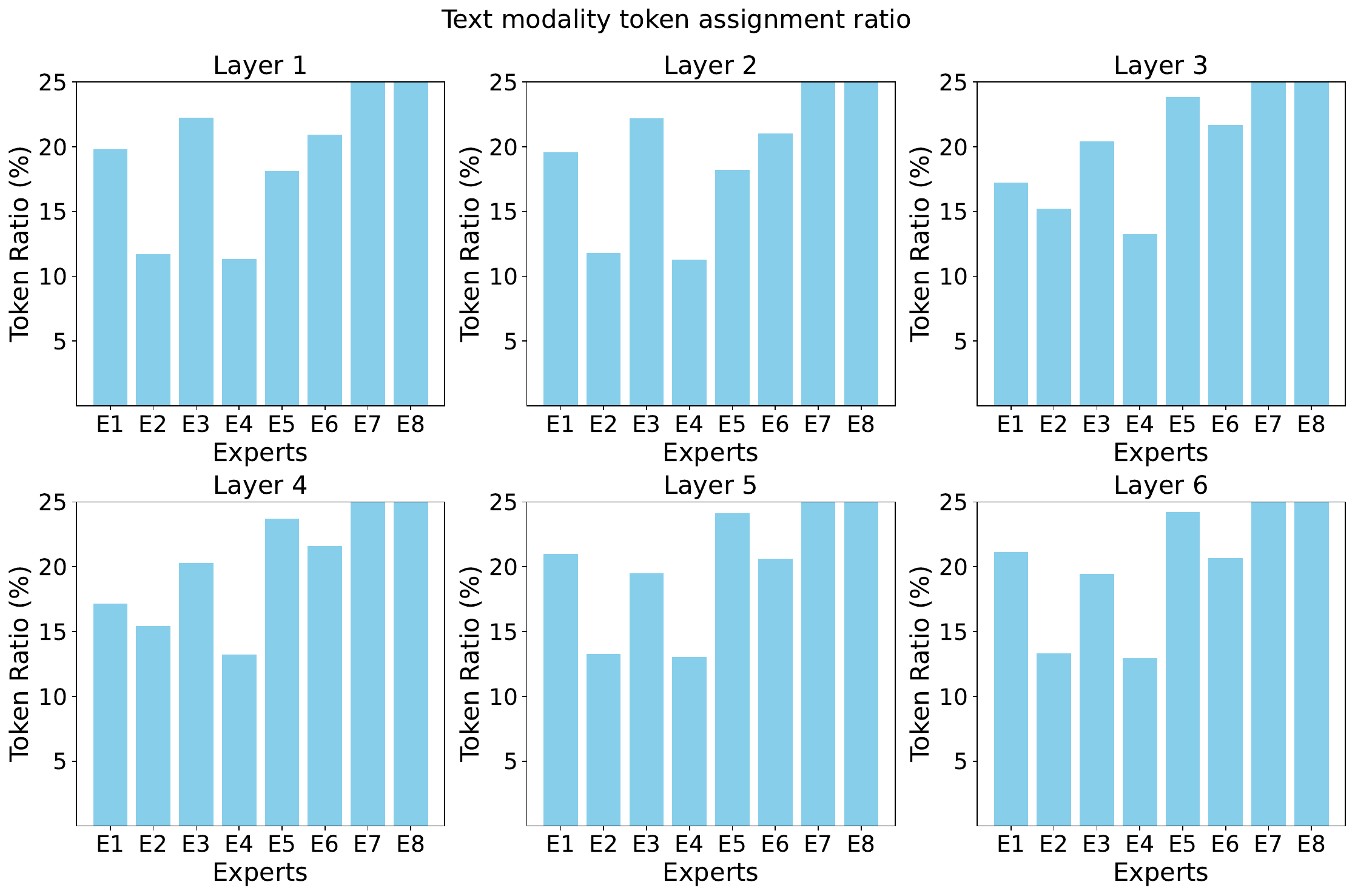}
    \end{subfigure}
    \vskip -0.1in
    \caption{Visualization of the token assignment ratios to each expert in each layer by the router on COCO — image tokens (top) and text tokens (bottom).}
    \label{fig:image-text-assign-ratio-coco}  
\end{figure}

The routing is balanced across all transformer layers. The average token ratios assigned to each expert for both text and image modalities on ImageNet and COCO are shown in Figure~\ref{fig:image-text-assign-ratio} and Figure~\ref{fig:image-text-assign-ratio-coco}, respectively.

Token assignment appears more balanced for images than for text, likely due to the higher number of tokens in images, many of which carry similar or redundant information. This redundancy facilitates a more uniform distribution of tokens across experts. In contrast, text is typically more discrete and information-dense, leading to a stronger preference for certain experts that specialize in specific linguistic patterns.

\newpage

\subsection{Expert preference pattern}

\begin{figure}[ht]  
    \centering  
    \begin{subfigure}
        \centering
        \includegraphics[width=1.0\linewidth]{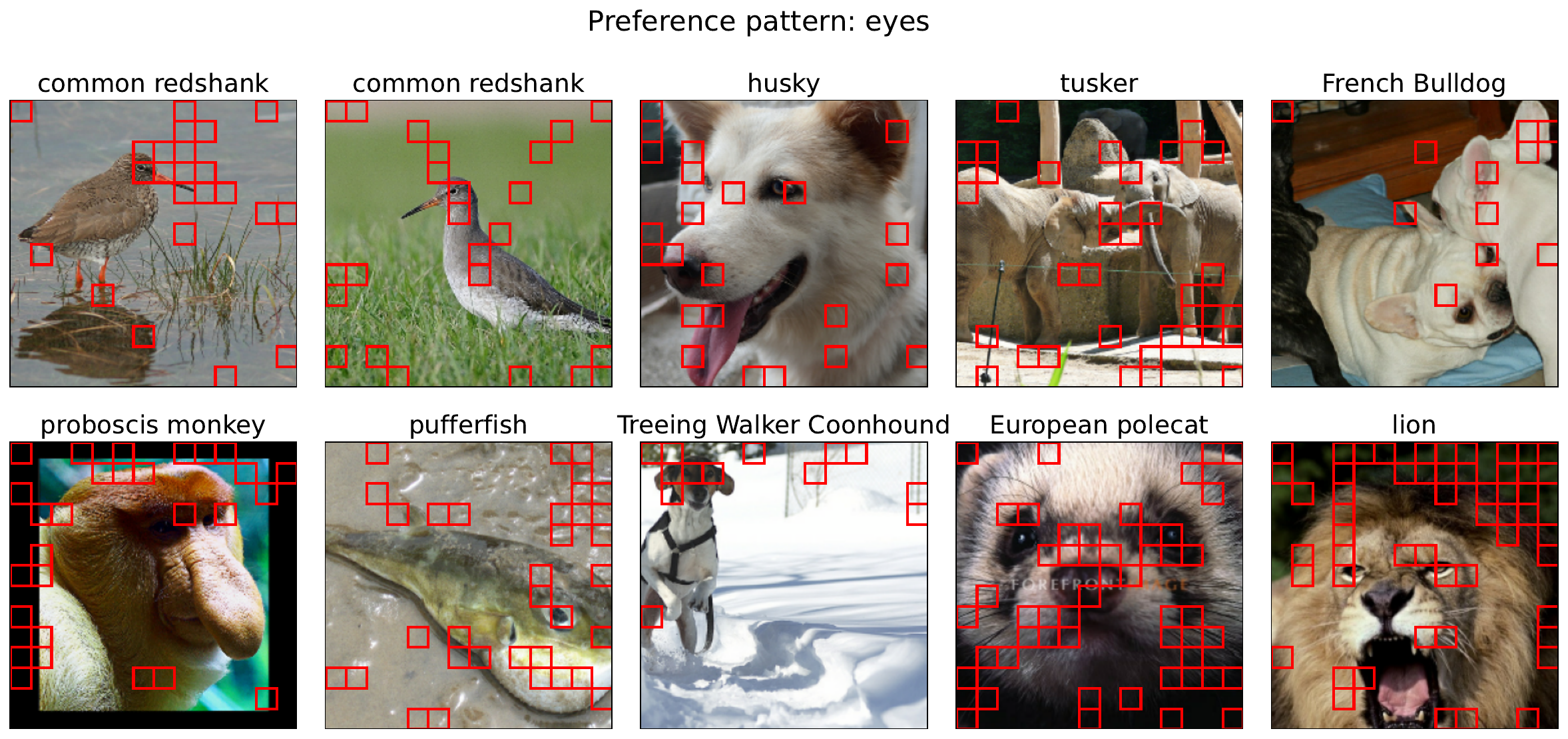}
    \end{subfigure}
    \begin{subfigure}
        \centering
        \includegraphics[width=1.0\linewidth]{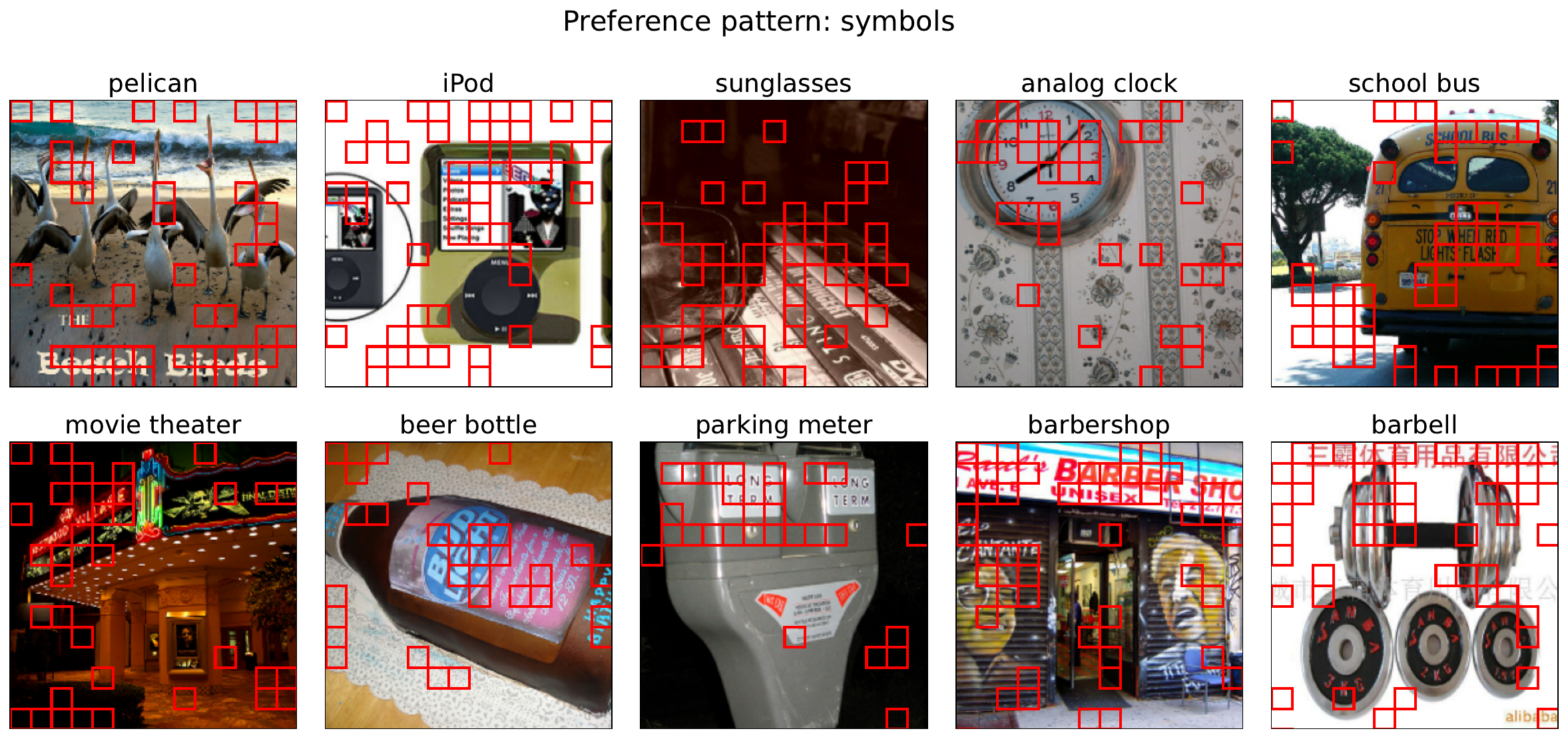}
    \end{subfigure}
    \vskip -0.1in
    \caption{Visualization of expert preference pattern examples. Red bounding boxes highlight the image tokens assigned to the expert.}
    \label{fig:expert-preference-pattern}  
\end{figure}

We observe distinct preference patterns among experts. As illustrated in Figure~\ref{fig:expert-preference-pattern}, one expert predominantly processes tokens related to "eyes," consistently attending to the eye regions of various animals and humans. Another expert demonstrates specialization in "symbols," handling text tokens from a wide range of contexts, including posters, machine interfaces, book titles, store signage, and instructional materials.

\end{document}

%% file: tabels/methodology-compare-no-limoe.tex
\begin{table}[t]
\caption{Ablation study comparing shared vs. separated backbones and training from scratch vs. sparse upcycling, evaluated on ImageNet (Accuracy@1 \%) and COCO/Flickr30K text-to-image (T2I)/image-to-text (I2T) retrieval (Recall@1 \%)}
\label{tab:methodology-compare-no-limoe}
\begin{center}
\begin{small}
\begin{sc}
\setlength{\tabcolsep}{1.2 pt}
\resizebox{1.0\linewidth}{!}{
\begin{tabular}{lc|ccccc}
\toprule
\multirow{3}{*}{Backbone} & \multirow{3}{*}{Upcycle?} & Imagenet &\multicolumn{2}{c}{COCO} &\multicolumn{2}{c}{Flickr30K} \\ 
& & & T2I &I2T &T2I &I2T \\
& & Acc@1 &R@1 &R@1 &R@1 &R@1 \\
\midrule
Shared & N & 69.7 &46.7 &65.6 &71.4 &86.3 \rule{0pt}{3ex} \\
Shared & Y & 75.2 &51.6 &\textbf{72.7} &78.0 &92.0 \rule{0pt}{3ex} \\
Separated & N & 74.5 &\textbf{53.1} &70.6 &78.3 &88.2 \rule{0pt}{3ex} \\
Separated & Y & \textbf{76.9} &52.1 &71.5 &\textbf{80.9} &\textbf{92.3} \rule{0pt}{3ex} \\
\bottomrule
\end{tabular}
}
\vskip -0.5in
\end{sc}
\end{small}
\end{center}
\end{table}

%% file: tabels/zero_shot_metrics_flops.tex
\begin{table*}[t]
\vskip -0.2in
\caption{Performance comparison of \name, dense models, and LIMOE across model sizes. \name is upcycled from a 440k-step CLIP checkpoint with 350k additional steps; a 790k-step dense CLIP is trained for fair comparison.
}
\label{zero-shot-metrics-flops-steps}
\begin{center}
\small
\begin{sc}
\setlength{\tabcolsep}{4pt}
\resizebox{0.85\textwidth}{!}{
\begin{tabular}{l|c|c|cc|cccc|cccc}
\toprule
\multirow{3}{*}{Model} &\multirow{3}{*}{Steps} &\multirow{3}{*}{Inference} &\multicolumn{2}{c|}{Imagenet} &\multicolumn{4}{c|}{COCO retrieval} &\multicolumn{4}{c}{Flickr30k retrieval} \\
& & & \multicolumn{2}{c|}{classification} &T2I &T2I &I2T &I2T &T2I &T2I &I2T &I2T \\
&(k) &GFLOPS &Acc@1 &Acc@5 &R@1 &R@5 &R@1 &R@5 &R@1 &R@5 &R@1 &R@5 \\
\midrule
\multicolumn{13}{c}{\bf \textit{B/32}} \\
\midrule
OpenAI-CLIP &- &14.8 &63.2 &88.8 &30.8 &55.9 &51.6 &75.7 &- &- &- &- \rule{0pt}{3ex} \\
LIMOE &- &22.3 &67.5 &- &31.0 &- &45.7 &- &- &- &- &- \rule{0pt}{3ex} \\
\cdashline{1-13}
CLIP (ours) &440 &14.8 &72.4 &92.8 &41.7 &68.2 &62.3 &84.3 &68.0 &90.0 &\textbf{86.5} &\textbf{97.7} \rule{0pt}{3ex} \\
CLIP (ours) &790 &14.8 &72.4 &92.8 &41.9 &67.8 &62.4 &84.1 &67.8 &88.7 &85.6 &96.4 \rule{0pt}{3ex} \\
\rowcolor{lightblue}
\name & 790 &19.6 &\textbf{73.2} &\textbf{93.3} &\textbf{47.3} &\textbf{74.0} &\textbf{66.6} &\textbf{86.7} &\textbf{72.9} &\textbf{91.9} &85.9 &96.9 \rule{0pt}{3ex} \\
\midrule
\multicolumn{13}{c}{\bf \textit{B/16}} \\
\midrule
OpenAI-CLIP &- &41.2 &68.4 &91.9 &33.1 &58.4 &53.8 &77.9 &- &- &- &- \rule{0pt}{3ex} \\
LIMOE &- &48.7 &73.7 &- &36.2 &- &51.3 &- &- &- &- &-\rule{0pt}{3ex} \\
\cdashline{1-13}
CLIP (ours) &440 &41.2 &76.0 &94.7 &44.4 &70.3 &65.7 &87.0 &73.6 &92.1 &88.0 &97.8 \rule{0pt}{3ex} \\
CLIP (ours) &790 &41.2 &76.8 &\textbf{95.1} &44.9 &70.8 &66.0 &86.6 &74.3 &92.7 &88.9 &98.0 \rule{0pt}{3ex} \\
\rowcolor{lightblue}
\name & 790 &54.3 &\textbf{76.9} &\textbf{95.1} &\textbf{52.1} &\textbf{77.6} &\textbf{71.5} &\textbf{89.2}  &\textbf{80.9} &\textbf{95.6} &\textbf{92.3} &\textbf{99.2} \rule{0pt}{3ex} \\
\midrule
\multicolumn{13}{c}{\bf \textit{L/14}} \\
\midrule
OpenAI-CLIP &- &175.5 &75.3 &94.5 &36.1 &60.8 &57.7 &79.1 &- &- &- &- \rule{0pt}{3ex} \\
\cdashline{1-13}
CLIP (ours) &440 &175.5 &81.1 &96.4 &49.6 &74.4 &70.9 &89.6 &78.4 &94.7 &91.9 &\textbf{99.2} \rule{0pt}{3ex} \\
CLIP (ours) &790 &175.5 &\textbf{81.6} &\textbf{96.6} &50.2 &75.2 &71.4 &89.9 &79.3 &94.9 &91.7 &99.0 \rule{0pt}{3ex} \\
\rowcolor{lightblue}
\name & 790 &231.7 &81.2 & \textbf{96.6} &\textbf{53.9} &\textbf{79.4} &\textbf{73.8} &\textbf{92.0}  &\textbf{82.0} &\textbf{96.1} &\textbf{92.4} &99.1 \rule{0pt}{3ex} \\
\bottomrule
\end{tabular}
}
\vskip -0.5in
\end{sc}
\end{center}
\end{table*}

%% file: tabels/appendix_hyper_parameters.tex
\begin{table}[h]
\caption{Training hyper-parameters and settings for dense CLIP used for sparse upcycling and \name}
\label{tab:hyper-parameter}
\vskip 0.15in
\begin{center}
\begin{small}
\begin{sc}
\renewcommand{\arraystretch}{1.5}
\resizebox{0.8\textwidth}{!}{
\begin{tabular}{lc}
\toprule
\multicolumn{2}{c}{\textbf{General}} \\
\midrule
Batch size & 32768 \\
\rowcolor{lightgray} 
Image size & $224 \times 224$ \\
Text tokenizer & T5~\cite{raffel2023exploringlimitstransferlearning}, lowercase \\
\rowcolor{lightgray} 
Text maximum length & 77 tokens \\
Optimizer & AdamW ($\beta_1 = 0.9, \beta_2 = 0.98$) \\
\rowcolor{lightgray} 
LR schedule & cosine decays with linear warm-up (first 2k steps) \\
Dropout rate & 0.0 \\
\midrule

\multicolumn{2}{c}{\textbf{MoE}} \\
\midrule
Inner structure & Pre-Layer Normalization~\cite{xiong2020layernormalizationtransformerarchitecture} \\
\rowcolor{lightgray} 
Router type & Top-2 routing \\
Expert capacity factor ($C$) & 2.0 (both text and image) \\
\rowcolor{lightgray} 
MoE position & [dense, sparse] (half of MLP layers replaced by MoE layers) \\
Load balance loss weight &0.01 \\
\rowcolor{lightgray} 
Router-z loss weight & 0.0001 \\
\midrule

\multicolumn{2}{c}{\textbf{Dense Model}} \\
\midrule
Steps & 439087 (\textit{i.e.,} $\sim$ 14B examples seen) \\
\rowcolor{lightgray} 
Peak learning rate (LR) & $5e^{-4}$ \\
Weight decay & 0.2 \\
\midrule

\multicolumn{2}{c}{\textbf{\name}} \\
\midrule
Steps & 351269 (\textit{i.e.,} $\sim$ 11B examples seen) \\
\rowcolor{lightgray} 
Peak learning rate (LR) & $5e^{-5}$ \\
Weight decay & 0.05 \\
\rowcolor{lightgray} 
Expert count & 8 (for text and image separately) \\
\midrule

\multicolumn{2}{c}{\textbf{Sparse Model}} \\
\midrule
Steps & 790356 (\textit{i.e.,} $\sim$ 25B examples seen) \\
\rowcolor{lightgray} 
Peak learning rate (LR) & $5e^{-4}$ \\
Weight decay & 0.2 \\
\rowcolor{lightgray} 
Expert count & 8 \\

\bottomrule
\end{tabular}}
\end{sc}
\end{small}
\end{center}
\end{table}

%% file: tabels/appendix_full_method_compare_with_LIMOE.tex


\begin{table*}[h]
\caption{All results from Table \ref{tab:methodology-compare-no-limoe} and Figure \ref{fig:limoe-compare} as discussed in Section \ref{comparison-methodology}}
\label{tab:appendix-recipe-study}
\begin{center}
\begin{small}
\begin{sc}
\begin{tabular}{lccccc}
\toprule
\multirow{2}{*}{Model} &\multicolumn{1}{c}{Imagenet} &\multicolumn{2}{c}{COCO} &\multicolumn{2}{c}{Flickr30K} \\
&Acc@1 &T2I R@1 &I2T R@1 &T2I R@1 &I2T R@1 \\
\midrule
\rowcolor{lightgray} 
Shared & 69.7 &46.7 &65.6 &71.4 &86.3 \rule{0pt}{3ex} \\
\quad+LIMOE Aux. Loss &73.1 &49.7 &69.7 &75.6 &87.9 \rule{0pt}{3ex} \\
\quad $\Delta$ & +3.4 & +3.0 & +4.1 & +4.2 & +1.6 \\
\cdashline{1-6}
\rowcolor{lightgray} 
Shared-UpCycle &75.2 &51.6 &\textbf{72.7} &78.0 &92.0 \rule{0pt}{3ex} \\
\quad+LIMOE Aux. Loss &73.9 &50.9 &70.1 &73.8 &85.5 \rule{0pt}{3ex} \\
\quad $\Delta$ & -1.3 & -0.7 & -2.6 & -4.2 & -6.5 \\
\cdashline{1-6}
\rowcolor{lightgray} 
Separated &74.5 &\textbf{53.1} &70.6 &78.3 &88.2 \rule{0pt}{3ex} \\
\quad+LIMOE Aux. Loss &72.6 &46.4 &62.6 &73.4 &85.2 \rule{0pt}{3ex} \\
\quad $\Delta$ & -2.0 & -6.7 & -8.0 & -4.9 & -3.0 \\
\cdashline{1-6}
\rowcolor{lightgray} 
Separated-UpCycle &\textbf{76.9} &52.1 &71.5 &\textbf{80.9} &\textbf{92.3} \rule{0pt}{3ex} \\
\quad+LIMOE Aux. Loss &75.9 &52.9 &73.5 &81.3 &92.5 \rule{0pt}{3ex} \\
\quad $\Delta$ & -1.0 & +0.8 & +2.0 & +0.4 & +0.2 \\
\bottomrule
\end{tabular}
\end{sc}
\end{small}
\end{center}
\end{table*}

%% file: tabels/appendix_imagenet_variants.tex
\begin{table*}[h]
\caption{Performance on ImageNet Variants of \name, dense models across model sizes.
}
\label{appendix_imagenet_variants}
\begin{center}
\small
\begin{sc}
\setlength{\tabcolsep}{8pt}
\begin{tabular}{l|c|cc|cc|cc}
\toprule
\multirow{3}{*}{Model} &\multirow{3}{*}{Steps} &\multicolumn{2}{c|}{Imagenet-V2} &\multicolumn{2}{c|}{Imagenet-A} &\multicolumn{2}{c}{Imagenet-R} \\
& & \multicolumn{2}{c|}{classification} &\multicolumn{2}{c|}{classification} &\multicolumn{2}{c}{classification} \\
&(k) &Acc@1 &Acc@5 &Acc@1 &Acc@5 &Acc@1 &Acc@5 \\
\midrule
\multicolumn{8}{c}{\bf \textit{B/32}} \\
\midrule
CLIP (ours) &440 &64.0 &88.0 &32.1 &65.4 &80.0 &92.7 \rule{0pt}{3ex} \\
CLIP (ours) &790 &64.0 &87.7 &32.6 &64.8 &\textbf{80.3} &92.7 \rule{0pt}{3ex} \\
\name & 790 &\textbf{65.1} &\textbf{88.5} &\textbf{34.2} &\textbf{66.8} &79.9 &\textbf{92.8} \rule{0pt}{3ex} \\
\midrule
\multicolumn{8}{c}{\bf \textit{B/16}} \\
\midrule
CLIP (ours) &440 &68.2 &90.7 &47.7 &78.5 &84.6 &95.3 \rule{0pt}{3ex} \\
CLIP (ours) &790 &69.3 &\textbf{91.2} &\textbf{50.5} &\textbf{80.2} &85.9 &\textbf{95.8} \rule{0pt}{3ex} \\
\name & 790 &\textbf{69.6} &91.1 &49.2 &78.6 &\textbf{85.9} &95.6 \rule{0pt}{3ex} \\
\midrule
\multicolumn{8}{c}{\bf \textit{L/14}} \\
\midrule
CLIP (ours) &440 &74.3 &93.5 &67.1 &\textbf{89.0} &90.6 &97.7 \rule{0pt}{3ex} \\
CLIP (ours) &790 &\textbf{74.7} &93.6 &\textbf{68.2} &88.9 &\textbf{91.2} &\textbf{97.8} \rule{0pt}{3ex} \\
\name & 790 &73.8 &\textbf{93.6} &66.1 &88.1 &89.7 &97.3 \rule{0pt}{3ex} \\
\bottomrule
\end{tabular}
\end{sc}
\end{center}
\end{table*}

%% file: tabels/appendix_modality_moe.tex
\begin{table*}[h]
\caption{All results from Figure \ref{fig:moe-modality}. MoE-text: MoE layers are added into text modality only. MoE-image: MoE layers are added into image modality only. MoE-both: MoE layers are added into both text and image modalities.}
\label{tab:appendix-modality}
\vskip 0.15in
\begin{center}
\begin{small}
\begin{sc}
\begin{tabular}{lcccccc}
\toprule
\multirow{3}{*}{Model} &\multirow{3}{*}{Steps} &\multicolumn{1}{c}{Imagenet} &\multicolumn{2}{c}{COCO} &\multicolumn{2}{c}{Flickr30K} \\
& &Acc@1 &T2I R@1 &I2T R@1 &T2I R@1 &I2T R@1 \\

\midrule
\multirow{5}{*}{MoE-text} 
&0 &70.2 &42.7 &52.5 &72.1 &79.4 \rule{0pt}{3ex} \\
&5 & 72.3 &42.2 &63.2 &70.0 &86.9 \rule{0pt}{3ex} \\
&100 & 73.3 &42.9 &63.1 &70.4 &88.0 \rule{0pt}{3ex} \\
&200 & 75.4 &42.8 &64.1 &72.7 &88.6 \rule{0pt}{3ex} \\
&350 & 77.2 &45.5 &66.0 &74.2 &89.6 \rule{0pt}{3ex} \\
\midrule
\multirow{5}{*}{MoE-image} &0 &62.6 &29.1 &50.9 &56.1 &73.7 \rule{0pt}{3ex} \\
&5 & 72.4 &41.4 &62.5 &70.0 &84.8 \rule{0pt}{3ex} \\
&100 & 73.8 &42.2 &63.6 &71.7 &88.2 \rule{0pt}{3ex} \\
&200 & 75.6 &43.8 &65.0 &72.2 &88.4 \rule{0pt}{3ex} \\
&350 & 77.6 &45.5 &66.3 &74.5 &89.4 \rule{0pt}{3ex} \\

\midrule
\multirow{5}{*}{MoE-both} &0 &57.2 &33.0 &45.3 &62.7 &73.4 \rule{0pt}{3ex} \\
&5 & 71.4 &44.9 &67.9 &71.6 &87.8 \rule{0pt}{3ex} \\
&100 & 72.8 &49.7 &71.4 &74.9 &89.1 \rule{0pt}{3ex} \\
&200 & 75.1 &49.0 &68.5 &73.1 &85.1 \rule{0pt}{3ex} \\
&350 & 76.9 &52.1 &71.5 &80.9 &92.3 \rule{0pt}{3ex} \\

\bottomrule
\end{tabular}
\end{sc}
\end{small}
\end{center}
\end{table*}

%% file: tabels/appendix_capacity_factor_imagenet_variants.tex
\begin{table*}[h]
\caption{All results from Figure \ref{fig:capacity-compare}}
\label{tab:appendix-capacity-compare-imagenet-variants}
\vskip 0.15in
\begin{center}
\begin{small}
\begin{sc}
\setlength{\tabcolsep}{1 pt}
\resizebox{0.98\textwidth}{!}{
\begin{tabular}{lcccccccc}
\toprule
\multirow{3}{*}{Model} &\multicolumn{1}{c}{Imagenet} &\multicolumn{1}{c}{Imagenet-V2} &\multicolumn{1}{c}{Imagenet-A} &\multicolumn{1}{c}{Imagenet-R} &\multicolumn{2}{c}{COCO} &\multicolumn{2}{c}{Flickr30K} \\
&Acc@1 &Acc@1 &Acc@1 &Acc@1 &T2I R@1 &I2T R@1 &T2I R@1 &I2T R@1 \\

\midrule
$C_{image}=2, C_{text}=2$ &76.9 &69.4 &49.8 &83.9 &\textbf{52.1} &\textbf{71.5} &\textbf{80.9} &\textbf{92.3} \rule{0pt}{3ex} \\
$C_{image}=4, C_{text}=2$ &\textbf{78.4} &\textbf{71.0} &\textbf{53.7} &\textbf{86.9} &46.3 &66.9 &75.5 &90.2 \rule{0pt}{3ex} \\
\bottomrule
\end{tabular}
}
\end{sc}
\end{small}
\end{center}
\end{table*}

%% file: acl_latex.bbl
\begin{thebibliography}{28}
\providecommand{\natexlab}[1]{#1}

\bibitem[{Cherti et~al.(2023)Cherti, Beaumont, Wightman, Wortsman, Ilharco, Gordon, Schuhmann, Schmidt, and Jitsev}]{Cherti_2023}
Mehdi Cherti, Romain Beaumont, Ross Wightman, Mitchell Wortsman, Gabriel Ilharco, Cade Gordon, Christoph Schuhmann, Ludwig Schmidt, and Jenia Jitsev. 2023.
\newblock \href {https://doi.org/10.1109/cvpr52729.2023.00276} {Reproducible scaling laws for contrastive language-image learning}.
\newblock In \emph{2023 IEEE/CVF Conference on Computer Vision and Pattern Recognition (CVPR)}. IEEE.

\bibitem[{Deng et~al.(2009)Deng, Dong, Socher, Li, Li, and Fei-Fei}]{deng2009imagenet}
Jia Deng, Wei Dong, Richard Socher, Li-Jia Li, Kai Li, and Li~Fei-Fei. 2009.
\newblock Imagenet: A large-scale hierarchical image database.
\newblock In \emph{2009 IEEE conference on computer vision and pattern recognition}, pages 248--255. Ieee.

\bibitem[{Du et~al.(2022)Du, Huang, Dai, Tong, Lepikhin, Xu, Krikun, Zhou, Yu, Firat, Zoph, Fedus, Bosma, Zhou, Wang, Wang, Webster, Pellat, Robinson, Meier-Hellstern, Duke, Dixon, Zhang, Le, Wu, Chen, and Cui}]{du2022glamefficientscalinglanguage}
Nan Du, Yanping Huang, Andrew~M. Dai, Simon Tong, Dmitry Lepikhin, Yuanzhong Xu, Maxim Krikun, Yanqi Zhou, Adams~Wei Yu, Orhan Firat, Barret Zoph, Liam Fedus, Maarten Bosma, Zongwei Zhou, Tao Wang, Yu~Emma Wang, Kellie Webster, Marie Pellat, Kevin Robinson, and 8 others. 2022.
\newblock \href {https://arxiv.org/abs/2112.06905} {Glam: Efficient scaling of language models with mixture-of-experts}.
\newblock \emph{Preprint}, arXiv:2112.06905.

\bibitem[{Fang et~al.(2023)Fang, Jose, Jain, Schmidt, Toshev, and Shankar}]{fang2023datafilteringnetworks}
Alex Fang, Albin~Madappally Jose, Amit Jain, Ludwig Schmidt, Alexander Toshev, and Vaishaal Shankar. 2023.
\newblock \href {https://arxiv.org/abs/2309.17425} {Data filtering networks}.
\newblock \emph{Preprint}, arXiv:2309.17425.

\bibitem[{Fedus et~al.(2022)Fedus, Zoph, and Shazeer}]{fedus2022switchtransformersscalingtrillion}
William Fedus, Barret Zoph, and Noam Shazeer. 2022.
\newblock \href {https://arxiv.org/abs/2101.03961} {Switch transformers: Scaling to trillion parameter models with simple and efficient sparsity}.
\newblock \emph{Preprint}, arXiv:2101.03961.

\bibitem[{Gan et~al.(2022)Gan, Li, Li, Wang, Liu, and Gao}]{gan2022visionlanguagepretrainingbasicsrecent}
Zhe Gan, Linjie Li, Chunyuan Li, Lijuan Wang, Zicheng Liu, and Jianfeng Gao. 2022.
\newblock \href {https://arxiv.org/abs/2210.09263} {Vision-language pre-training: Basics, recent advances, and future trends}.
\newblock \emph{Preprint}, arXiv:2210.09263.

\bibitem[{Hendrycks et~al.(2021{\natexlab{a}})Hendrycks, Basart, Mu, Kadavath, Wang, Dorundo, Desai, Zhu, Parajuli, Guo, Song, Steinhardt, and Gilmer}]{hendrycks2021facesrobustnesscriticalanalysis}
Dan Hendrycks, Steven Basart, Norman Mu, Saurav Kadavath, Frank Wang, Evan Dorundo, Rahul Desai, Tyler Zhu, Samyak Parajuli, Mike Guo, Dawn Song, Jacob Steinhardt, and Justin Gilmer. 2021{\natexlab{a}}.
\newblock \href {https://arxiv.org/abs/2006.16241} {The many faces of robustness: A critical analysis of out-of-distribution generalization}.
\newblock \emph{Preprint}, arXiv:2006.16241.

\bibitem[{Hendrycks et~al.(2021{\natexlab{b}})Hendrycks, Zhao, Basart, Steinhardt, and Song}]{hendrycks2021naturaladversarialexamples}
Dan Hendrycks, Kevin Zhao, Steven Basart, Jacob Steinhardt, and Dawn Song. 2021{\natexlab{b}}.
\newblock \href {https://arxiv.org/abs/1907.07174} {Natural adversarial examples}.
\newblock \emph{Preprint}, arXiv:1907.07174.

\bibitem[{Jia et~al.(2021)Jia, Yang, Xia, Chen, Parekh, Pham, Le, Sung, Li, and Duerig}]{jia2021scaling}
Chao Jia, Yinfei Yang, Ye~Xia, Yi-Ting Chen, Zarana Parekh, Hieu Pham, Quoc Le, Yun-Hsuan Sung, Zhen Li, and Tom Duerig. 2021.
\newblock Scaling up visual and vision-language representation learning with noisy text supervision.
\newblock In \emph{International conference on machine learning}, pages 4904--4916. PMLR.

\bibitem[{Komatsuzaki et~al.(2023)Komatsuzaki, Puigcerver, Lee-Thorp, Ruiz, Mustafa, Ainslie, Tay, Dehghani, and Houlsby}]{komatsuzaki2023sparseupcyclingtrainingmixtureofexperts}
Aran Komatsuzaki, Joan Puigcerver, James Lee-Thorp, Carlos~Riquelme Ruiz, Basil Mustafa, Joshua Ainslie, Yi~Tay, Mostafa Dehghani, and Neil Houlsby. 2023.
\newblock \href {https://arxiv.org/abs/2212.05055} {Sparse upcycling: Training mixture-of-experts from dense checkpoints}.
\newblock \emph{Preprint}, arXiv:2212.05055.

\bibitem[{Lin et~al.(2014)Lin, Maire, Belongie, Hays, Perona, Ramanan, Doll{\'a}r, and Zitnick}]{lin2014microsoft}
Tsung-Yi Lin, Michael Maire, Serge Belongie, James Hays, Pietro Perona, Deva Ramanan, Piotr Doll{\'a}r, and C~Lawrence Zitnick. 2014.
\newblock Microsoft coco: Common objects in context.
\newblock In \emph{Computer Vision--ECCV 2014: 13th European Conference, Zurich, Switzerland, September 6-12, 2014, Proceedings, Part V 13}, pages 740--755. Springer.

\bibitem[{Liu et~al.(2023)Liu, Li, Wu, and Lee}]{liu2023visualinstructiontuning}
Haotian Liu, Chunyuan Li, Qingyang Wu, and Yong~Jae Lee. 2023.
\newblock \href {https://arxiv.org/abs/2304.08485} {Visual instruction tuning}.
\newblock \emph{Preprint}, arXiv:2304.08485.

\bibitem[{Mustafa et~al.(2022)Mustafa, Riquelme, Puigcerver, Jenatton, and Houlsby}]{mustafa2022multimodalcontrastivelearninglimoe}
Basil Mustafa, Carlos Riquelme, Joan Puigcerver, Rodolphe Jenatton, and Neil Houlsby. 2022.
\newblock \href {https://arxiv.org/abs/2206.02770} {Multimodal contrastive learning with limoe: the language-image mixture of experts}.
\newblock \emph{Preprint}, arXiv:2206.02770.

\bibitem[{Plummer et~al.(2017)Plummer, Wang, Cervantes, Caicedo, Hockenmaier, and Lazebnik}]{flickr30k}
Bryan~A. Plummer, Liwei Wang, Chris~M. Cervantes, Juan~C. Caicedo, Julia Hockenmaier, and Svetlana Lazebnik. 2017.
\newblock \href {https://doi.org/10.1007/s11263-016-0965-7} {Flickr30k entities: Collecting region-to-phrase correspondences for richer image-to-sentence models}.
\newblock \emph{Int. J. Comput. Vision}, 123(1):74–93.

\bibitem[{Radford et~al.(2021)Radford, Kim, Hallacy, Ramesh, Goh, Agarwal, Sastry, Askell, Mishkin, Clark, Krueger, and Sutskever}]{radford2021learningtransferablevisualmodels}
Alec Radford, Jong~Wook Kim, Chris Hallacy, Aditya Ramesh, Gabriel Goh, Sandhini Agarwal, Girish Sastry, Amanda Askell, Pamela Mishkin, Jack Clark, Gretchen Krueger, and Ilya Sutskever. 2021.
\newblock \href {https://arxiv.org/abs/2103.00020} {Learning transferable visual models from natural language supervision}.
\newblock \emph{Preprint}, arXiv:2103.00020.

\bibitem[{Raffel et~al.(2023)Raffel, Shazeer, Roberts, Lee, Narang, Matena, Zhou, Li, and Liu}]{raffel2023exploringlimitstransferlearning}
Colin Raffel, Noam Shazeer, Adam Roberts, Katherine Lee, Sharan Narang, Michael Matena, Yanqi Zhou, Wei Li, and Peter~J. Liu. 2023.
\newblock \href {https://arxiv.org/abs/1910.10683} {Exploring the limits of transfer learning with a unified text-to-text transformer}.
\newblock \emph{Preprint}, arXiv:1910.10683.

\bibitem[{Ramesh et~al.(2021)Ramesh, Pavlov, Goh, Gray, Voss, Radford, Chen, and Sutskever}]{ramesh2021zeroshottexttoimagegeneration}
Aditya Ramesh, Mikhail Pavlov, Gabriel Goh, Scott Gray, Chelsea Voss, Alec Radford, Mark Chen, and Ilya Sutskever. 2021.
\newblock \href {https://arxiv.org/abs/2102.12092} {Zero-shot text-to-image generation}.
\newblock \emph{Preprint}, arXiv:2102.12092.

\bibitem[{Rao et~al.(2022)Rao, Zhao, Chen, Tang, Zhu, Huang, Zhou, and Lu}]{rao2022densecliplanguageguideddenseprediction}
Yongming Rao, Wenliang Zhao, Guangyi Chen, Yansong Tang, Zheng Zhu, Guan Huang, Jie Zhou, and Jiwen Lu. 2022.
\newblock \href {https://arxiv.org/abs/2112.01518} {Denseclip: Language-guided dense prediction with context-aware prompting}.
\newblock \emph{Preprint}, arXiv:2112.01518.

\bibitem[{Recht et~al.(2019)Recht, Roelofs, Schmidt, and Shankar}]{recht2019imagenetclassifiersgeneralizeimagenet}
Benjamin Recht, Rebecca Roelofs, Ludwig Schmidt, and Vaishaal Shankar. 2019.
\newblock \href {https://arxiv.org/abs/1902.10811} {Do imagenet classifiers generalize to imagenet?}
\newblock \emph{Preprint}, arXiv:1902.10811.

\bibitem[{Shankar et~al.(2020)Shankar, Roelofs, Mania, Fang, Recht, and Schmidt}]{pmlr-v119-shankar20c}
Vaishaal Shankar, Rebecca Roelofs, Horia Mania, Alex Fang, Benjamin Recht, and Ludwig Schmidt. 2020.
\newblock Evaluating machine accuracy on {I}mage{N}et.
\newblock In \emph{Proceedings of the 37th International Conference on Machine Learning}, volume 119 of \emph{Proceedings of Machine Learning Research}, pages 8634--8644. PMLR.

\bibitem[{Shazeer et~al.(2017)Shazeer, Mirhoseini, Maziarz, Davis, Le, Hinton, and Dean}]{shazeer2017outrageouslylargeneuralnetworks}
Noam Shazeer, Azalia Mirhoseini, Krzysztof Maziarz, Andy Davis, Quoc Le, Geoffrey Hinton, and Jeff Dean. 2017.
\newblock \href {https://arxiv.org/abs/1701.06538} {Outrageously large neural networks: The sparsely-gated mixture-of-experts layer}.
\newblock \emph{Preprint}, arXiv:1701.06538.

\bibitem[{Wu et~al.(2024)Wu, Timofeev, Chen, Zhang, Duan, Liu, Zheng, Shlens, Du, Gan, and Yang}]{wu2024mofilearningimagerepresentations}
Wentao Wu, Aleksei Timofeev, Chen Chen, Bowen Zhang, Kun Duan, Shuangning Liu, Yantao Zheng, Jonathon Shlens, Xianzhi Du, Zhe Gan, and Yinfei Yang. 2024.
\newblock \href {https://arxiv.org/abs/2306.07952} {Mofi: Learning image representations from noisy entity annotated images}.
\newblock \emph{Preprint}, arXiv:2306.07952.

\bibitem[{Xiong et~al.(2020)Xiong, Yang, He, Zheng, Zheng, Xing, Zhang, Lan, Wang, and Liu}]{xiong2020layernormalizationtransformerarchitecture}
Ruibin Xiong, Yunchang Yang, Di~He, Kai Zheng, Shuxin Zheng, Chen Xing, Huishuai Zhang, Yanyan Lan, Liwei Wang, and Tie-Yan Liu. 2020.
\newblock \href {https://arxiv.org/abs/2002.04745} {On layer normalization in the transformer architecture}.
\newblock \emph{Preprint}, arXiv:2002.04745.

\bibitem[{Xue et~al.(2024)Xue, Zheng, Fu, Ni, Zheng, Zhou, and You}]{xue2024openmoeearlyeffortopen}
Fuzhao Xue, Zian Zheng, Yao Fu, Jinjie Ni, Zangwei Zheng, Wangchunshu Zhou, and Yang You. 2024.
\newblock \href {https://arxiv.org/abs/2402.01739} {Openmoe: An early effort on open mixture-of-experts language models}.
\newblock \emph{Preprint}, arXiv:2402.01739.

\bibitem[{Zeng et~al.(2024)Zeng, Guo, Fei, Yin, Zhou, Li, Sun, Yan, Lin, and Qiu}]{zeng2024turnwasteworthrectifying}
Zhiyuan Zeng, Qipeng Guo, Zhaoye Fei, Zhangyue Yin, Yunhua Zhou, Linyang Li, Tianxiang Sun, Hang Yan, Dahua Lin, and Xipeng Qiu. 2024.
\newblock \href {https://arxiv.org/abs/2402.12399} {Turn waste into worth: Rectifying top-$k$ router of moe}.
\newblock \emph{Preprint}, arXiv:2402.12399.

\bibitem[{Zhang et~al.(2024)Zhang, Qu, Zhu, and Cheng}]{zhang2024clipmoebuildingmixtureexperts}
Jihai Zhang, Xiaoye Qu, Tong Zhu, and Yu~Cheng. 2024.
\newblock \href {https://arxiv.org/abs/2409.19291} {Clip-moe: Towards building mixture of experts for clip with diversified multiplet upcycling}.
\newblock \emph{Preprint}, arXiv:2409.19291.

\bibitem[{Zhou et~al.(2022)Zhou, Yang, Loy, and Liu}]{Zhou_2022}
Kaiyang Zhou, Jingkang Yang, Chen~Change Loy, and Ziwei Liu. 2022.
\newblock \href {https://doi.org/10.1007/s11263-022-01653-1} {Learning to prompt for vision-language models}.
\newblock \emph{International Journal of Computer Vision}, 130(9):2337–2348.

\bibitem[{Zoph et~al.(2022)Zoph, Bello, Kumar, Du, Huang, Dean, Shazeer, and Fedus}]{zoph2022stmoedesigningstabletransferable}
Barret Zoph, Irwan Bello, Sameer Kumar, Nan Du, Yanping Huang, Jeff Dean, Noam Shazeer, and William Fedus. 2022.
\newblock \href {https://arxiv.org/abs/2202.08906} {St-moe: Designing stable and transferable sparse expert models}.
\newblock \emph{Preprint}, arXiv:2202.08906.

\end{thebibliography}
